\documentclass{article}

\usepackage{graphicx}
\usepackage{subcaption}
\usepackage{booktabs}
\usepackage[normalem]{ulem}

\usepackage{xcolor}
\usepackage{listings}
\lstdefinelanguage{json}{
  basicstyle=\ttfamily\scriptsize,
  numbers=none,
  showstringspaces=false,
  breaklines=true,
  frame=single,
  rulecolor=\color{black},
  stringstyle=\color{black},
  morestring=[b]",
}

\usepackage{iclr2026_conference,times}
\iclrfinalcopy
\usepackage[utf8]{inputenc} 
\usepackage[T1]{fontenc}    
\usepackage{hyperref}       
\usepackage{url}            
\usepackage{booktabs}       
\usepackage{amsfonts}       
\usepackage{nicefrac}       
\usepackage{microtype}      
\usepackage{xcolor}         

\usepackage{amsmath}

\usepackage{tikz}
\usepackage{varwidth}
\usetikzlibrary{positioning}
\usepackage{fancyvrb}

\usepackage{algorithm}
\usepackage{algpseudocode}

\newcommand{\bx}{{\bf x}}
\newcommand{\bz}{{\bf z}}
\newcommand{\by}{{\bf y}}
\newcommand{\bc}{{\bf c}}
\newcommand{\bu}{{\bf u}}
\newcommand{\bq}{{\bf q}}
\newcommand{\bw}{{\bf w}}
\newcommand{\bY}{{\bf Y}}
\newcommand{\bX}{{\bf X}}
\newcommand{\bU}{{\bf U}}
\newcommand{\bS}{{\bf S}}
\newcommand{\bV}{{\bf V}}
\newcommand{\bR}{{\bf R}}

\newcommand{\bLambda}{{\bf \Lambda}}

\newcommand{\bsig}{{\bf \sigma}}

\DeclareMathOperator{\diag}{diag}
\DeclareMathOperator{\tr}{tr}

\title{GenZ: Foundational models as latent variable generators within traditional statistical models}

\author{%
  Marko Jojic \\
  Arizona State University, Tempe, AZ, USA
  \And
  Nebojsa Jojic \\
  Microsoft Research, Redmond, WA, USA
}

\begin{document}

\maketitle

\begin{abstract}
We present GenZ, a hybrid model that bridges foundational models and statistical modeling through interpretable semantic features. While large language models possess broad domain knowledge, they often fail to capture dataset-specific patterns critical for prediction tasks. Our approach addresses this by discovering semantic feature descriptions through an iterative process that contrasts groups of items identified via statistical modeling errors, rather than relying solely on the foundational model's domain understanding. We formulate this as a generalized EM algorithm that jointly optimizes semantic feature descriptors and statistical model parameters. The method prompts a frozen foundational model to classify items based on discovered features, treating these judgments as noisy observations of latent binary features that predict real-valued targets through learned statistical relationships. We demonstrate the approach on two domains: house price prediction (hedonic regression) and cold-start collaborative filtering for movie recommendations. On house prices, our model achieves 12\% median relative error using discovered semantic features from multimodal listing data, substantially outperforming a GPT-5 baseline (38\% error) that relies on the LLM's general domain knowledge. For Netflix movie embeddings, our model predicts collaborative filtering representations with 0.59 cosine similarity purely from semantic descriptions---matching the performance that would require approximately 4000 user ratings through traditional collaborative filtering. The discovered features reveal dataset-specific patterns (e.g., architectural details predicting local housing markets, franchise membership predicting user preferences) that diverge from the model's domain knowledge alone.
\end{abstract}

\section{Introduction}
\label{sec:intro}

Given an item's semantics, such as a text description, image, or audio recording, large foundational models can answer specific questions about the content, and thus convert the item's semantic representation into a canonical discrete representation. Such a representation is useful for traditional statistical models—such as graphical generative models, belief networks, or probabilistic circuits—which can capture statistics from external, possibly proprietary, datasets related to these items.

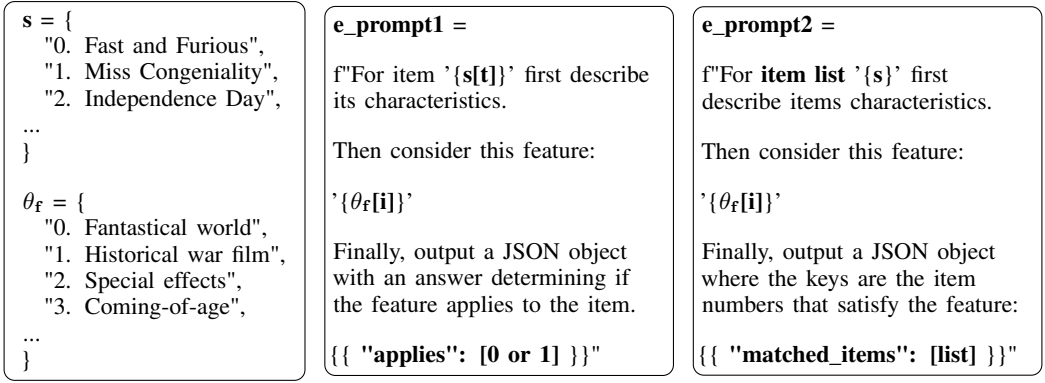
\begin{figure}[t]
\centering
\begin{tikzpicture}[
    box/.style={draw, rounded corners, minimum width=4cm, minimum height=2.3cm, align=left, text width=3.5cm},
    promptbox/.style={draw, rounded corners, minimum width=4cm, minimum height=2.5cm, align=left, text width=4.4cm},
    node distance=0.5cm
]

\node[box] (box1) {
    \begin{varwidth}{\linewidth}
    \begin{small}
\begin{flushleft}
\textbf{s} = \{

\phantom{aa}"0. Fast and Furious",

\phantom{aa}"1. Miss Congeniality",

\phantom{aa}"2. Independence Day",

...

\}

\vspace{1em}
\textbf{$\bf\theta_f$} = \{

\phantom{aa}"0. Fantastical world",

\phantom{aa}"1. Historical war film",

\phantom{aa}"2. Special effects",

\phantom{aa}"3. Coming-of-age",

...

\}

\end{flushleft}
\end{small}
    \end{varwidth}
};

\node[promptbox, right=0.25cm of box1] (box2) {
    \begin{varwidth}{\linewidth}
    
\begin{small}
\begin{flushleft}
\textbf{e\_prompt1} =

\vspace{1em}

f"For  item '\{\textbf{s[t]}\}' first describe its characteristics.
\vspace{1em}

Then consider this feature:
\vspace{1em}

'\{$\bf\theta_f$\textbf{[i]}\}'
\vspace{1em}

Finally, output a JSON object with an answer determining if the feature applies to the item.
\vspace{1em}

\{\{ \textbf{"applies": [0 or 1]} \}\}"

\end{flushleft}
\end{small}
    \end{varwidth}
};

\node[promptbox, right=0.25cm of box2] (box3){
    \begin{varwidth}{\linewidth}
        
\begin{small}
\begin{flushleft}
\textbf{e\_prompt2} =

\vspace{1em}

f"For {\textbf {item list}} '\{\textbf{s}\}' first describe items characteristics.
\vspace{1em}

Then consider this feature:
\vspace{1em}

'\{$\bf\theta_f$\textbf{[i]}\}'
\vspace{1em}

Finally, output a JSON object where the keys are the item numbers that satisfy the feature:
\vspace{1em}

\{\{ \textbf{"matched\_items": [list]} \}\}"

\end{flushleft}
\end{small}
    \end{varwidth}
};

\end{tikzpicture}
\caption{Feature classification $h(s,\theta_f)$ with extraction prompt strings shown in Python syntax. Items $s$ in the list are featurized using feature descriptions in $\theta_f$. The first prompt is used within a double loop ($t,i$) over items and features. The second is looped only over $i$ (and batches of items \{s\}) as each call retrieves the indices of items for which $z_i = 1$ (with $z_i = 0$ for all other items).}
\label{fig:prompts}
\end{figure}

Movies are an example of such items. We can use an LLM to featurize them as binary vectors based on answers to questions as illustrated in Fig. \ref{fig:prompts}. These binary vectors can help explain the patterns found in the movie-user ratings matrix. Collaborative filtering techniques reduce the observation matrix to embeddings, which capture usage statistics absent from the foundational models' training data. The foundational models, however, can reason about the semantic content—such as the movie synopsis, critical response, credits, and other metadata—either from their training or by retrieving information from a database as part of a RAG system.

We describe how appropriate prompts can be learned from such item statistics by jointly modeling the semantic features and the statistical patterns in the data.

We study a general hybrid statistical-foundational model defined as follows. Let $s$ be the semantic representation of an item, e.g., text with a movie title or its description. Given a collection of semantic feature descriptors $\theta_f$, such as 'historical war film', a foundational model, which we refer to as function $h$, makes a judgment whether the feature applies to the item (Figure \ref{fig:prompts}). Let $\bz$ be the true binary feature vector that allows for possible errors in either the model's judgment or the wrongly worded feature descriptors. We model uncertainty by defining a conditional distribution $p(\bz|s,\theta_f, \theta_e)$, where $\bz=[z_1, z_2, ..., z_{n_f}]^T$ is a discrete feature vector (binary, in the example in the figure and in our experiments), $\theta_f$ is a collection of semantic feature descriptors, such as 'historical war film', and $\theta_e=[p^e_1, p^e_2,..., p^e_{n_f}]$ is a set of estimated probabilities of feature classification errors. A feature vector is then linked to observation $\by$ through a classical statistical model $p(\by|\bz,\theta_y)$, possibly involving latent variables or some other way of modeling structure, and which has its own parameters $\theta_y$. By jointly optimizing for the parameters of the model $p(\by|s)=\sum_{\bz}p(\by|\bz)p(\bz|s)$ based on a large number of pairs $(\by,s)$ we can discover the feature descriptions, or \textbf{the right questions to ask}, in order to explain the statistics in the dataset. More generally, hybrid models could have more than one category of items with a (learnable) statistical model of their relationships, for example geographic regions and native plants.

The idea of using interpretable intermediate representations to bridge inputs and predictions has a long history in machine learning. Concept bottleneck models (CBM) \cite{koh2020concept, kim2023probabilistic} formalized this approach by training neural networks to predict \emph{pre-defined} human-interpretable concepts from inputs, then using those concepts to make final predictions. Recent work has leveraged large language models (LLMs) to overcome the need for manually labeled concepts, e.g., \cite{benara2024craftinginterpretableembeddingsasking,feng2025bayesianconceptbottleneckmodels,chattopadhyay2023variationalinformationpursuitinterpretable,zhong2025explainingdatasetswordsstatistical,ludan2024interpretablebydesigntextunderstandingiteratively}. These methods may or may not use the CBM formulation but share with it the goal of feature interpretability. In addition, they have a goal of feature \emph{discovery}, which is also one of the goals here. They address the difficulty of modeling and jointly adjusting the semantic and statistical parts of the model—in our notation $p(\bz|s)$ and $p(\by|\bz)$—in different ways. 

In most cases, the target $\by$ is a discrete variable (the goal is a classification through intermediate concepts), and thus the classification model $p(\by|\bz)$ is a simple logistic regressor or an exhaustive conditional table \cite{chattopadhyay2023variationalinformationpursuitinterpretable,feng2025bayesianconceptbottleneckmodels, zhong2025explainingdatasetswordsstatistical}. The optimization of the featurizer $p(\bz|s)$ primarily relies on the LLM's domain understanding (such as a task description), but some methods also provide examples of misclassified items, e.g., \cite{zhong2025explainingdatasetswordsstatistical, ludan2024interpretablebydesigntextunderstandingiteratively}. 

When real-valued multi-dimensional targets $\by$ are modeled, showing items with large prediction errors is much less informative because each item may have a different error vector, and thus item-driven discovery is usually abandoned. For example, in \cite{benara2024craftinginterpretableembeddingsasking} multidimensional targets (fMRI recordings) are modeled, and the method relies on direct LLM prompting about potentially predictive features based on domain knowledge alone, followed by feature selection, rather than showing specific items to contrast.

A fundamental limitation shared by these approaches is their reliance on the LLM's domain understanding to propose discriminative features. This works well when the LLM has been trained on relevant data and the task aligns with common patterns in its training distribution. However, as our experiments demonstrate, domain understanding alone is insufficient when dataset-specific patterns diverge from general knowledge---such as house pricing dynamics in particular markets, or user preferences captured in collaborative filtering from specific time periods or demographics. Moreover, for high-dimensional real-valued targets $\by$, individual prediction errors have multidimensional structure that cannot be easily communicated to an LLM by simply showing "incorrect" examples.

\begin{figure}[t]
\centering
\begin{tikzpicture}[
    box/.style={draw, rounded corners, minimum width=4cm, minimum height=2.5cm, align=left, text width=3.2cm},
    promptbox/.style={draw, rounded corners, minimum width=4cm, minimum height=2.5cm, align=left, text width=8cm},
    node distance=0.5cm
]

\node[box] (box1) {
    \begin{varwidth}{\linewidth}
    \begin{small}
\begin{flushleft}
\textbf{Positive} = \{

\phantom{aa}"Armageddon",

\phantom{aa}"The Rock",

\phantom{aa}"Lethal Weapon 4",

\phantom{aa}"Con Air",

\phantom{aa}"Twister"

\}

\vspace{1em}
\textbf{Negative} = \{

\phantom{aa}"Mystic River",

\phantom{aa}"Lost in Translation",

\phantom{aa}"Collateral",

\phantom{aa}"Fahrenheit 9/11",

\phantom{aa}"Big Fish"

\}

\end{flushleft}
\end{small}
    \end{varwidth}
};

\node[promptbox, right=0.5cm of box1] (box2) {
    \begin{varwidth}{\linewidth}
    
\begin{small}
\begin{flushleft}
\textbf{m\_prompt} =

\vspace{1em}

f"Consider the two groups and first describe the items and their characteristics.
\vspace{1em}

Positive:
'\{Positive\}'
\vspace{1em}

Negative:
'\{Negative\}'
\vspace{1em}

What characteristic/feature separates the positive group? Think step by step then output the answer as a JSON object
\vspace{1em}

\{\{  "characteristic": [text] \}\}"

\end{flushleft}
\end{small}
    \end{varwidth}
};

\end{tikzpicture}

\caption{Discovering feature descriptions for prompts in Fig. \ref{fig:prompts}. 
\label{sec:prompt_mining}
When the statistical model identifies a binary split in the data, we can prompt a foundational model to generate an explanation by providing example items from each group. This explanation can then be used as a feature descriptor in $\theta_f$. The prompt need not include all members of each group, only representative examples.}
\label{fig:prompt_mining}
\end{figure}
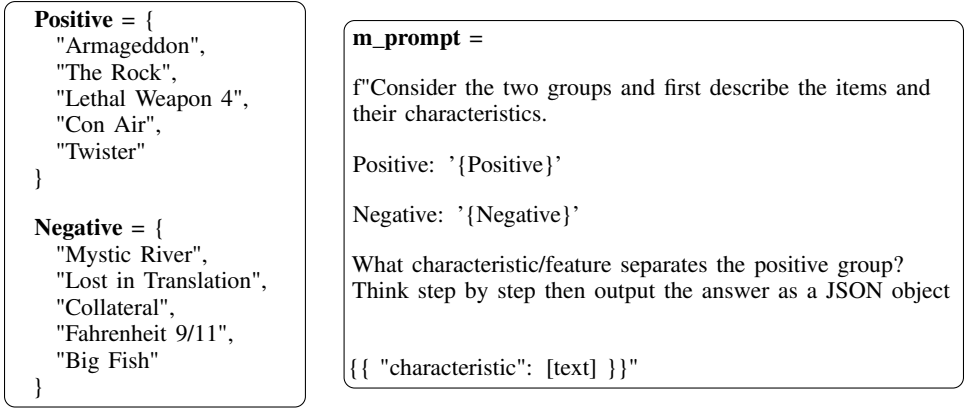

Our approach differs in several key aspects. First, as in some of the recent work, we avoid concept supervision by using a frozen foundational model as an oracle within $p(\bz|s)$, but with trainable uncertainty parameters $\theta_e$. Second, our feature mining is driven by contrasting groups of items identified through the statistical model's posterior $q(\bz)$ rather than by querying the LLM's domain knowledge or showing individual misclassified examples (Figure \ref{fig:prompt_mining}). This group-based contrast allows the method to discover dataset-specific patterns that may not align with the LLM's training distribution—or even in cases where the LLM has no understanding of what the target $\by$ represents. (We do, however, require the (M)LLM to reason about the semantic items $s$, either based on knowledge encoded in its weights or by retrieving information from a database as in RAG systems). Third, we support arbitrary mappings $p(\by|\bz)$ including nonlinear functions of binary features for high-dimensional real-valued observations, shifting the burden of modeling complex feature interactions from the LLM to the statistical model. The LLM is only responsible for explaining the semantic coherence of binary splits discovered based on the model's prediction errors, making it applicable to domains where the relationship between semantics and observations is not well understood in advance.

We refer to our hybrid model as GenZ (or GenAIz?) as it relies on a pre-trained generative AI model to generate candidate latents $\bz$ in a prediction model $s \rightarrow \bz \rightarrow \by$. We derive a generalized EM algorithm for training it, which naturally requires investigation of items $s$ when tuning parameters $\theta_f$ of the featurizer $p(\bz|s)$.

\section{Semantic model}
\label{sec:semantic_model}
Before developing the full algorithm, we first describe the semantic part: the semantic parameters $\theta_f$, how $p(\bz|s)$ depends on them and on a foundational model, and how statistical signals from the model enable updating these parameters in an iterative EM-like algorithm.

\subsection{Extracting semantic features: Modeling \texorpdfstring{$p(\bz|s)$}{p(z|s)}}

The foundational model's reasoning abilities are sufficient to featurize a semantic item using standard classification prompts. A generic version of this is illustrated in Fig. \ref{fig:prompts}. In the example shown, the items are represented with text, in this case movie titles that large language models are likely familiar with. Each string in the list is an item description $s$—however, the text describing each item could be expanded to include more details about the movie, or a RAG system with access to a movie database could be used in place of an LLM. We use the output $h(s,\theta_f)$ of such systems cautiously: we associate an uncertainty level, i.e., a probability $p^e_j$ of mislabeling $h \ne z$ for entry $z_j$ in $\bz$. Assuming that $\bz$ is not deterministically linked to $s$ makes it a \textbf{latent} variable in the $p(\by,\bz|s)$ model, allowing flexibility in inference and enabling updates to the semantic feature descriptions in $\theta_f$: If the posterior over $\by$ can override the foundational model's determination of whether an item satisfies the feature, this provides a signal for updating the feature semantics.

\subsection{Mining semantic features}

As already alluded to, the feature semantics $\theta_f$ are \textbf{not} hand-crafted but learned from the dataset statistics. The statistical signal from the data indicates how the items should be separated along a particular feature $j$, i.e., how the $z_j$ for different items can be altered so that the model explains the data better. In our setup, we deal with binary features, so the signal separates the items into two groups, and we seek a semantic explanation for this separation, $\theta_f$.

Specifically, for fixed model parameters, focusing on a particular binary feature $z_i \in\{0,1\}$ for one semantic item $s$, the prior and posterior distributions are generally different: 
\begin{equation}
    p(z_i|s) \neq p(z_i|\by,s)=\sum_{\{z_j\}_{j\neq i}}p(z_i| \by, \{z_j\}_{j\neq i}, s)  \nonumber
\end{equation}

Thus, sampling the posterior distribution (referred to below as $q$) for several different items $s^t$ yields new feature assignments $z_i^t$. These help update the feature descriptor ${\theta_f}_i$ used in $p(z_i|s,\theta_f)$ by using the prompt in Fig. \ref{fig:prompt_mining}. There, the positive group has a few items $s^t$ for which $z_i^t=1$ and the negative group has a few items for which $z_i^t=0$. Furthermore, we can select group exemplars to contain both semantic items where the posterior and the prior agree, and where they disagree. This encourages small refinements, e.g., from "fearless female characters in an action movie" to "strong female leads".

\section{Statistical model}

Given an item's semantics $s$, e.g., a movie description, and its corresponding observed vector $\by$, e.g., the movie's embedding derived from the statistics in the movie-user matrix, the log likelihood can be bounded as
\begin{eqnarray}
    \log p(\by|s) \geq B&=&\sum_\bz q(\bz)\log \frac{p(\by|\bz)p(\bz|s)}{q(\bz)} \nonumber\\
    &=& E_q[\log p(\by|\bz)]+E_q[\log p(\bz|s)]-E_q[\log q(\bz)],
    \label{eq:bound}
\end{eqnarray}
where $E_q$ refers to an expectation under the distribution $q$. Depending on the variational approximation of $q$ and the model $p(\by|\bz)$ these expectations can be computed exactly, as illustrated by the example of the mean-field $q$ and linear $p(\by|\bz)$ in Section \ref{sec:linear_observation_model}. Otherwise, a sampling method can be used, often with very few samples due to the iterative nature of the learning algorithm. In experiments (Section \ref{sec:experiments}), we simply evaluate at the mode of $q$.

The bound is tight when the variational distribution $q$ is equal to the true posterior $p(\bz|\by,s)$, otherwise $q(\bz)$ serves as an approximate posterior. We refer to the three additive terms as the observation part $T_1=E_q[\log p(\by|\bz)]$, the feature part $T_2=E_q[\log p(\bz|s)]$, and the posterior entropy $T_3=-E_q[\log q(\bz)]$.

Given a dataset of pairs $\{s^t,\by^t\}_{t=1}^T$, the optimization criterion—the bound on the log likelihood of the entire dataset—is
\begin{equation}
\label{eq:approx_LL}
    L(\{q^t\},\theta_f, \theta_e, \theta_y)=\sum_t B_t = \sum_t  E_{q^t}[\log p(\by^t|\bz)]+E_{q^t}[\log p(\bz|s)]-E_{q^t}[\log q^t(\bz)] 
\end{equation}

In general, maximizing the criterion $L$ w.r.t. all variables except semantic model parameters (feature descriptions) $\theta_f$ is straightforward: we can use existing packages or derive an EM algorithm specific to the choice of the model's statistical distributions. The variational distributions $q^t(\bz)$ depend only on the individual bounds for each data pair $(\by^t, s^t)$ in (\ref{eq:bound}) and can be fitted independently for given model parameters $\theta_f, \theta_e, \theta_y$.

The (approximate) posteriors $q_t(z_i)$ provide the statistical signal discussed in Section \ref{sec:prompt_mining} for tuning the semantic feature descriptions ${\theta_f}_i$: sampling them separates the samples $s^t$ into two groups, and exemplars from each are used to prompt the foundational model with the prompt in Fig. \ref{fig:prompt_mining} for a new definition of the $i$-th feature. The optimization of $L$ then consists of steps that improve it w.r.t. the statistical model parameters and the (approximate) posteriors $q$, alternating with updates of the feature descriptions using those posteriors.

This general recipe can be used in a variety of ways, with different choices for the structure of the conditionals in the model $p$ and different approximations of the posterior $q$ leading to a variety of algorithms including variational learning, belief propagation and sampling \cite{frey-jojic-pami-tutorial}. We now develop specific modeling choices and a learning algorithm we use in our experiments.

{\bf Binary feature model}

As discussed, the feature model uses a prompt-based feature classification, where each feature is treated independently based on its description ${\theta_f}_i$ (Fig. \ref{fig:prompts})
\begin{equation}
    p(\bz|s,\theta_f, \theta_e)=\prod_i p(z_i|s, {\theta_f}_i, p^e_i),
\end{equation}
with the uncertainty (probability of error) $p^e_i$, so that
\begin{equation}
    p(z_i|s)=(1-p^e_i)^{[h(s,{\theta_f}_i)=z_i]}(p^e_i)^{[h(s,{\theta_f}_i)\neq z_i]}
\end{equation}
where $h$ is the binary function that uses one of the prompts in Fig. \ref{fig:prompts} to detect the feature ${\theta_f}_i$ in the semantic item $s$. The expression above uses the indicator function $[]$ to express that with the probability of error $p_e^i$, the feature indicator $z_i$ is different from $h(s,{\theta_f}_i)$. Denoting $h_i=h(s,{\theta_f}_i)$, we write
\begin{eqnarray}
\label{eq:feature_model}
    &\log p(z_i|s)=[z_i=1]\ell^1_i+[z_i=0]\ell^0_i, \\
    &\qquad \ell^1_i=[h_i=1]\log(1-p^e_i)+[h_i=0]\log p^e_i, \nonumber\\ 
    &\quad \ell^0_i=[h_i=1]\log p^e_i +[h_i=0]\log (1-p^e_i) \nonumber
\end{eqnarray}

Using a factorized posterior $q=\prod_i q(z_i)$, the feature term $T_2$ simplifies to 
\begin{eqnarray}
\label{eq:T2}
    T_2=E_q[\log p(\bz|s)]=\sum_i \sum_{z_i}q(z_i)\log p(z_i|s)=\sum_i q_i\ell^1_i+(1-q_i)\ell^0_i,
\end{eqnarray}
with $\ell^1_i$ and $\ell^0_i$ defined in (\ref{eq:feature_model}). Finally, the entropy term becomes
\begin{equation}
\label{eq:T3}
    T_3=-E_q[\log q(\bz)]=-\sum_i q_i\log q_i + (1-q_i)\log(1-q_i),  \quad q_i=q(z_i=1)
\end{equation}

{\bf General observation model}

We can soften an arbitrary mapping function $\by=f(\bz; \theta_m)$ by defining a distribution:
\begin{eqnarray}
    p(\by|\bz, \theta_y=\{\theta_m, \bsig^2_y\})&=&{\cal N}(\by| f(\bz, \theta_m), \Sigma=diag(\bsig^2_y)), \\
    T_1=E_q[\log p(\by|\bz)]&=&E_q\big[-\frac{1}{2}(\by-f(\bz))^T\Sigma^{-1}(y-f(\bz))\big]-\frac{1}{2}\log 2\pi|\Sigma|,
\end{eqnarray}
with $\theta_m$ being the mapping parameters of function $f$, e.g., regression weights in $\by={\bf w}^T\bz$, and $\bsig^2_y$ being a vector of variances, one for each dimension of $\by$. Different specialized inference algorithms for specific forms of function $f$ can be derived. For example, see the appendix for a procedure derived for a linear model (factor analysis). However, in our experiments, we use a generic inference technique using $f$ in plug-and-play fashion, described below.

{\bf Inference of $q(z_i)$ in the fully factorized (mean field) $q$ model}

Keeping all but one $q(z_i)$ fixed, we derive the update for the $i$-th feature's posterior by setting the derivative of the bound $B$ wrt parameters $q(z_i=1)$ and $q(z_i=0)$, subject to $q(z_i=1)+q(z_i=0)=1$. We can express the result as:
\begin{eqnarray}
    q(z_i=1) &\propto& e^{E_{q_{+i}}[\log p(\by|\bz)] +E_{q_{+i}}[\log p(\bz|s)] - E_{q_{+i}}[\log q(\bz)]}, \nonumber\\
    q(z_i=0) &\propto& e^{E_{q_{-i}}[\log p(\by|\bz)] +E_{q_{-i}}[\log p(\bz|s)] - E_{q_{-i}}[\log q(\bz)]}
\end{eqnarray}
where $q_{+i}=[z_i=1]\prod_{j\ne i} q(z_j)$, and $q_{+i}=[z_i=0]\prod_{j\ne i} q(z_j)$, i.e the "+i" and "-i" posteriors have their $z_i$ value clamped to 1 and 0 respectively in the factorized $q=\prod q(z_i)$. Upon normalization, as the entropy terms $E_{\bq^{+i}}[\log q(\bz)]$ and $E_{\bq^{-i}}[\log q(\bz)]$ cancel each other, we obtain:
\begin{equation}
\label{eq:q_i}
    q_i=q(z_i=1)=1-q(z_i=0)=\frac{1}{1+e^{E_{\bq^{-i}}[\log p(\by|\bz)]-E_{\bq^{+i}}[\log p(\by|\bz)]+\ell^0_i-\ell^1_i}}
\end{equation}
Interpretation: Whether or not $z_i$ should take value 1 or 0 for  particular semantic item $s$ depends on:
\begin{itemize}
    \item the assignment $h$ that the foundational model assigns it and the level of trust we have in that assignment (the probability of error $p^e_i$). The two are captured in prior log likelihoods $\ell_1, \ell_0$.
    \item error vector $\by-f(\bz)$,  scaled by the currently estimated variances $\bsig^2_y$ for different elements of the target $\by$,  for two possible assignments $z_i=1$ and $z_i=0$, with other assignments $z_j$ following the current distributions over other features $q(z_j)$
\end{itemize}
While expectations $E_{q_{+i}}$ and $E_{q_{+i}}$ can potentially be better estimated through sampling, in our experiments, we simply compute $\log p(\by|\bz)$ at the mode (assuming most likely values for other $z_j$ variables).

{\bf Optimizing numerical parameters of $p(\by | \bz)$ and $p(\bz|s)$}

Optimizing the bound $B$ wrt to parameters $\theta_m$ of the mapping in $\by\approx f(\bz; \theta_m)$ and the remaining uncertainty/variance $\bsig^2_y$ reduces to maximizing sum of $T_1$ terms over the traing set, i.e. $\sum_t E_q^t[\log p(\by^t|\bz^t)]$. Again, while sampling of $q$ might be helpful, in our experiments we simply use the mode and iterate:
\begin{eqnarray}
\label{eq:py}
    \theta_m &=& \arg \min \sum_t (\by^t-f(\bz^t; \theta_m))^T\Sigma^{-1}(\by^t-f(\bz^t; \theta_m)) \\
    \bsig^2_y &=& \frac{1}{T} \sum_t (\by^t-f(\bz^t; \theta_m))\circ (\by^t-f(\bz^t; \theta_m))
\end{eqnarray}
where $\bz^t$ is the binary vector of assignments of features $z_i$, either a sample from $q_t$ or at the mode of the posterior $q_t$ for the $t$-th out of $T$ entries in the data set of pairs $s_t, \by_t$. $\circ$ is a point-wise multiplication.

The semantic model has numeric parameters $p^e_i$, modeling the hybrid model's uncertainty in semantic function $h$ that makes an API call to the foundational model. By optimizing the bound wrt these parameters we obtain the update:
\begin{equation}
\label{eq:pe}
    p^e_i =\frac{1}{T}\sum_{t=1}^T q_i^t[h_i^t=0]+(1-q_i^t)[h_i^t=1]
\end{equation}

Thus, for a given set of semantic feature descriptors $\theta_f$, iterating equations (\ref{eq:q_i}) for each item $t$, and (\ref{eq:py}, \ref{eq:pe}) using all items in the training set, would fit the numerical model parameters and create the soft feature assignments $q^t$ that best balance those feature descriptors with the target real and possible multi-dimensional targets $\by^t$.

\section{Algorithms}
In section \ref{sec:prompt_mining} we pointed out that individual $q(z_i)$ distributions carry information about how the data should be split into two groups to create the mining prompt in Fig. \ref{fig:prompt_mining}. But as each of these distributions depend on the others, there is the question of good initialization. In our experiment we use a method akin to the variational EM view of boosting \cite{neal1998boosting}, albeit in our case applied to learning to predict real-valued multi-dimensional targets, rather than just for classification.

Note that if we iterate (\ref{eq:q_i}, \ref{eq:py}, \ref{eq:pe}) while keeping $p^e_i=0.5$ for a select feature $i$ the entire hybrid model and the posterior  $q(z_i)$ are decoupled from the foundational model's prediction $h_i$ which depends on feature description ${\theta_f}_i$ to provide its guess at $z_i$. Therefore, the iterative process is allowed to come up with the assignment of $z^t_i$ for each item  $t$ in such a way that maximizes the likelihood (and minimizes the prediction error). This provides a natural \emph{binary split} of the data into two groups based on error vectors. In case of predicting house prices (one dimensional $y$, those two groups would be the lower and higher priced houses relative to current prediction, but in case of multi-dimensional observations, such as movie embeddings, the split would be into two clusters, each with their own multi-dimensional adjustment to the current prediction.

A simple algorithm for adding features is shown in Algorithm \ref{alg:add_feature}. The key idea is to fit a model with only features $j \in [1..i]$ while keeping the new feature $i$ decoupled from the LLM prediction by fixing $p^e_i=0.5$. This allows the data to determine the optimal split before discovering its semantic explanation.

The algorithm operates on the model $p(\{\by^t, \{z_j^t\}_{j=1}^i, s^t\}_{t=1}^T)$, where features $j > i$ do not participate. By keeping $p^e_i=0.5$ fixed during the inner loop, the posterior $q_i^t$ is determined purely by how well $z_i^t$ helps explain $\by^t$, independent of any LLM prediction. This provides the statistical signal needed to discover meaningful features through the mining prompt.

As we show in experiments on the simple task of discovering binary representation of numbers, iteratively running Algorithm \ref{alg:add_feature} with increasing $i$ adds new features in coarse-to-fine manner, from those that split the data based on large differences in target $\by$ to those that contribute to explaining increasingly smaller differences.

{\bf Conditional (targeted) feature addition}

A possible problem with such coarse-to-fine approach, especially for nonlinear and easy to overfit functions $f(\bz)$, is that items with different combinations of features can be arbitrarily split with the feature $z_i$ we are updating. For example, suppose we are working on the house pricing dataset and are running the Algorithm \ref{alg:add_feature} for the second time, having added only one feature so far: $\theta_{f_1}=$\emph{'Located in Arizona desert communities'}. Then as we iteratively look for a natural split $q_2$ it is possible that for items with high $q_1$, the procedure may split the data along a different semantic axis (e.g., presence of swimming pools) than for the items with low $q_1$. To deal with this, we alter Algorithm \ref{alg:add_feature} very slightly: When we sample positive and negative examples, we take them from items that have other features identical. This is akin to using the structured variational approximation $q=q(z_i|\{z_j\}_{j\ne i})\prod_{j\ne i} q(z_j)$, but instead of fitting this for all combinations of other features, we target only one combination at a time. The discovered feature $\theta_{f_i}$ is still used to assign values $h_i$ for all data samples, whether they were in the targeted group or not, so that it can be used in the next step.

To select the combination of features $\{z_j\}_{j=1}^{i-1}$ on which we condition the $i$-th feature discovery, we start with the cost being optimized in (\ref{eq:py}), from the observation part ($T_1$) of the approximate log likelihood based on a sample (or the mode) $\bz^t$ for each item:

\begin{equation}
\sum_t \big(\by^t-f(\bz^t)\big)^T\Sigma^{-1}\big(\by^t-f(\bz^t)\big),
\end{equation}
or equivalently, by partitioning over all observed combinations $\bc$ of the other features:
\begin{equation}
\label{eq:T1_by_combination}
\sum_{\bc} \sum_{t: \bz_{-i}^t=\bc} \big(\by^t-f(\bz^t)\big)^T\Sigma^{-1}\big(\by^t-f(\bz^t)\big) = \sum_{\bc} T_1^{\bc},
\end{equation}
where $T_1^{\bc}$ are the summands that show the total error due to each observed combination $\bc$ of the other features $\bz_{-i} = [z_1, \ldots, z_{i-1}]^T$. We select the combination $\bc^*$ with the largest total error $\bc^* = \arg\max_{\bc} T_1^{\bc}$. Algorithm \ref{alg:add_feature_conditional} shows the conditional (targeted) feature addition procedure. The key difference from Algorithm \ref{alg:add_feature} is that positive and negative examples are sampled from a targeted subgroup $G^*$ (items sharing the same values for features $1$ through $i-1$), selected by identifying which combination has the largest total error. The discovered feature $\theta_{f_i}$ is then applied to all items.

These algorithms add a feature but can also be used to \emph{update} a feature by removing it from the set and then adding a new one. While we can select a feature to update at random, we use a more efficient approach: we choose the feature whose removal affects the prediction (likelihood bound) the least, as shown in Algorithm \ref{alg:prune_feature}. A single feature update becomes a removal of one feature (Algorithm \ref{alg:prune_feature}) followed by adding a feature (Algorithm \ref{alg:add_feature}). Generally, we alternate an arbitrary number of feature adding steps with feature removal steps, as shown in Algorithm \ref{alg:expand_contract}.

\begin{algorithm}[H]
\caption{Basic feature discovery (non-targeted)}
\label{alg:add_feature}
\begin{algorithmic}[1]
\State \textbf{Input:} Dataset $\mathcal{D} = \{(\by^t, s^t)\}_{t=1}^T$, feature descriptors $\theta_{f_j}$, $j \in [1..i-1]$,  $n_{\text{inner}}$
\State \textbf{Output:} New feature descriptor $\theta_{f_i}$ and assignments $h_i^t$ for all items

\State // \emph{Initialize: Decouple new feature from LLM prediction}
\State Set $p^e_i \gets 0.5$ \Comment{Maximum uncertainty - feature decoupled from $h_i$}
\State Initialize $q_i^t \gets 0.5$ for all $t \in [1..T]$ \Comment{Uniform prior}

\State // \emph{Iterate with $p^e_i=0.5$ fixed: Optimize $q$ and $\theta_y$ on partial model}
\For{iteration $= 1$ to $n_{\text{inner}}$}
    \State // \emph{E-step: Update posteriors for all items and features}
    \For{each item $t = 1$ to $T$ and feature $j\in[1..i-1]$}
        \State Update $q_j^t \gets $ using (\ref{eq:q_i}) \Comment{Features $j \in [1..i]$ participate}
        \State Sample or take mode: $z_j^t \sim q_j^t$ or $z_j^t = [q_j^t > 0.5]$
    \EndFor
    
    \State // \emph{M-step: Update model parameters based on all posteriors}
    \State Update mapping parameters $\theta_m$ using (\ref{eq:py}) \Comment{Fit $f(\bz)$ with $\bz = [z_1, \ldots, z_i]^T$}
    \State Update observation variance $\bsig^2_y$ using (\ref{eq:py})
    \State Update error rates $p^e_j$ for $j < i$ using (\ref{eq:pe}) \Comment{But keep $p^e_i = 0.5$}
\EndFor

\State // \emph{Mining: Discover semantic explanation}
\State Sample $q_i^t$ to form groups:
\State \quad $\text{Pos} \gets \{s^t : q_i^t > \tau_{\text{high}}\}$ \Comment{Positive examples (up to $n_e$ items)}
\State \quad $\text{Neg} \gets \{s^t : q_i^t < \tau_{\text{low}}\}$ \Comment{Negative examples (up to $n_e$ items)}
\State $\theta_{f_i} \gets \text{LLM}(\text{Pos}, \text{Neg})$ \Comment{Mining prompt, Fig. \ref{fig:prompt_mining}}
\State Apply $\theta_{f_i}$ to all items: $h_i^t \gets p(z_i = 1 | s^t, \theta_{f_i})$ for $t=1,...,T$ \Comment{Extraction prompt, Fig. \ref{fig:prompts}}

\State \Return $\theta_{f_i}, \{h_i^t\}_{t=1}^T$
\end{algorithmic}
\end{algorithm}

\begin{algorithm}[H]
\caption{Conditional (Targeted) Feature Addition}
\label{alg:add_feature_conditional}
\begin{algorithmic}[1]
\State \textbf{Input:} Dataset $\mathcal{D} = \{(\by^t, s^t)\}_{t=1}^T$, feature descriptors $\theta_{f_j}$, $j \in [1..i-1]$, $n_{\text{inner}}$
\State \textbf{Output:} New feature descriptor $\theta_{f_i}$ and assignments $h_i^t$ for all items
\Statex
\Statex \emph{Lines 3--17: Same as Algorithm \ref{alg:add_feature} (initialize, iterate to obtain $q_i^t$ and $\bz^t$)}
\Statex
\setcounter{ALG@line}{17}

\State // \emph{Mining: Select targeted group and discover semantic explanation}
\State Select $\bc^* = \arg\max_{\bc} T_1^{\bc}$ where $T_1^{\bc} = \sum_{t: \bz_{-i}^t=\bc} (\by^t-f(\bz^t))^T\Sigma^{-1}(\by^t-f(\bz^t))$ \Comment{Eq. (\ref{eq:T1_by_combination})}

\State // \emph{Sample positive and negative groups from targeted combination}
\State $G^* \gets \{t : \bz_{-i}^t = \bc^*\}$
\State $\text{Pos} \gets \{s^t : t \in G^*, q_i^t > \tau_{\text{high}}\}$ \Comment{Positive examples from $G^*$ only}
\State $\text{Neg} \gets \{s^t : t \in G^*, q_i^t < \tau_{\text{low}}\}$ \Comment{Negative examples from $G^*$ only}
\State $\theta_{f_i} \gets \text{LLM}(\text{Pos}, \text{Neg})$ \Comment{Mining prompt, Fig. \ref{fig:prompt_mining}}
\State Apply $\theta_{f_i}$ to all items: $h_i^t \gets p(z_i = 1 | s^t, \theta_{f_i})$ for $t=1,...,T$ \Comment{Extraction prompt, Fig. \ref{fig:prompts}}

\State \Return $\theta_{f_i}, \{h_i^t\}_{t=1}^T$
\end{algorithmic}
\end{algorithm}

\vspace{1em}

\begin{algorithm}[H]
\caption{Prune Feature (remove least useful feature)}
\label{alg:prune_feature}
\begin{algorithmic}[1]
\State \textbf{Input:} Dataset $\mathcal{D} = \{(\by^t, s^t)\}_{t=1}^T$, feature descriptors $\{\theta_{f_j}\}_{j=1}^{n_f}$, model parameters $\theta_m, \bsig^2_y, \{p^e_j\}$
\State \textbf{Output:} Index $j^*$ of feature to remove, updated model with feature removed

\State // \emph{Compute current likelihood bound}
\State $L_{\text{full}} \gets \sum_{t=1}^T B^t$ using current model with all $n_f$ features \Comment{Equation (\ref{eq:bound})}

\State // \emph{Find feature with minimum impact on likelihood}
\State $\Delta L_{\min} \gets +\infty$, $j^* \gets \text{null}$
\For{each feature $j = 1$ to $n_f$}
    \State Create temporary model excluding feature $j$: $\{\theta_{f_k}\}_{k \neq j}$
    \State Set $h_k^t$ for all $t$ using extraction prompt for $k \neq j$ \Comment{Fig. \ref{fig:prompts}}
    
    \State // \emph{Refit model without feature $j$}
    \For{iteration $= 1$ to $n_{\text{refit}}$}
        \For{each item $t$ and feature $k \neq j$}
            \State Update $q_k^t$ using (\ref{eq:q_i})
        \EndFor
        \State Update $\theta_m, \bsig^2_y, \{p^e_k\}_{k \neq j}$ using (\ref{eq:py}, \ref{eq:pe})
    \EndFor
    
    \State // \emph{Compute likelihood without feature $j$}
    \State $L_{-j} \gets \sum_{t=1}^T B^t$ using model without feature $j$
    \State $\Delta L_j \gets L_{\text{full}} - L_{-j}$ \Comment{Impact of removing feature $j$}
    
    \If{$\Delta L_j < \Delta L_{\min}$}
        \State $\Delta L_{\min} \gets \Delta L_j$, $j^* \gets j$
    \EndIf
\EndFor

\State // \emph{Remove least useful feature}
\State Remove $\theta_{f_{j^*}}$ from feature set
\State Update model: reindex features $j > j^*$ and refit parameters

\State \Return $j^*$, updated model with $n_f - 1$ features
\end{algorithmic}
\end{algorithm}

\begin{algorithm}[H]
\caption{Expand-Contract (iterative feature refinement)}
\label{alg:expand_contract}
\begin{algorithmic}[1]
\State \textbf{Input:} Dataset $\mathcal{D} = \{(\by^t, s^t)\}_{t=1}^T$, $k_{\text{add}}$ (features to add), $k_{\text{cut}}$ (features to prune), max cycles
\State \textbf{Output:} Refined feature set $\{\theta_{f_j}\}$ and model parameters

\State Initialize with empty feature set or existing features
\For{cycle $= 1$ to max\_cycles}
    \State // \emph{Expansion phase: Add $k_{\text{add}}$ features}
    \For{$a = 1$ to $k_{\text{add}}$}
        \State $\theta_{f_{n_f+1}}, \{h_{n_f+1}^t\} \gets$ \textbf{Add Feature} (Algorithm \ref{alg:add_feature})
        \State $n_f \gets n_f + 1$
    \EndFor
    
    \State // \emph{Contraction phase: Remove $k_{\text{cut}}$ features}
    \For{$c = 1$ to $k_{\text{cut}}$}
        \State $j^* \gets$ \textbf{Prune Feature} (Algorithm \ref{alg:prune_feature})
        \State $n_f \gets n_f - 1$
    \EndFor
    
    \State // \emph{Check convergence}
    \If{feature set unchanged or likelihood improvement $< \epsilon$}
        \State \textbf{break}
    \EndIf
\EndFor

\State \Return Final feature set $\{\theta_{f_j}\}_{j=1}^{n_f}$ and model parameters
\end{algorithmic}
\end{algorithm}

\section{Experiments}
\label{sec:experiments}
Especially exciting applications of hybrid models discussed above involve situations in which the foundational model is capable of recognizing patterns in the data, i.e., features that some items share and others do not, but the high-dimensional target $\by$ is not understood well by the model, e.g., in scientific discovery based on new experiments. However, there is value in analyzing situations where we (humans) understand the links so that we can better understand what the model is discovering. This is especially interesting when the data involves well understood semantic items, but also novel (or dated!), or otherwise skewed statistics between these items and the targets $\by$ which we want to discover, despite a potential domain shift w.r.t. what the foundational model was trained on.

We analyzed three datasets we expected to be illustrative:
\begin{itemize}
    \item Learning binary: A toy dataset with items $s$ being strings $"0", "1", \ldots ,  "511"$, and the corresponding targets $y = \text{str2num}(s)$. We demonstrate that iterating Algorithm \ref{alg:add_feature} recovers a 9-bit binary representation.
    \item Hedonic regression on house prices: A multimodal dataset from \cite{ahmed2016house}, consisting of four listing images and metadata, which we turned into semantic items $s$ describing all information about a house and the target log price $y$ (Fig. \ref{fig:house-semantic-item}).
    \item Cold start in recommendation systems: The Netflix prize dataset. Singular value decomposition of the user-movie matrix is used to make 32-dimensional embedding $\by$ for each movie, and the associated semantic item is the movie title and the year of release.
\end{itemize}

Since LLMs can understand similarities between movies or houses and prices, there was a chance that the models could discover relevant features without our iterative algorithms. In fact, as discussed in Section \ref{sec:intro}, most prior work on using off-the-shelf LLMs to discover features useful in classification and, especially, regression focused on probing LLMs with the description of the task, rather than the data itself, and subsequent feature selection. Our baseline that represents such approaches, which we refer to as \emph{0-shot}, is to provide an LLM with an example list of semantic items, describe the task, and ask for around 50 yes/no characteristics that in the LLM's judgment would be most useful for the task.

Except for the first experiment, whose purpose is to illustrate coarse-to-fine nature of feature discovery with a linear model, we tested two $f(\bz)$ functions (softened in $p(\by|\bz)$ through learned variances $\sigma^2_y$): a linear model and a neural network with two hidden layers (64 and 32 units with ReLU activations and dropout rate 0.1), trained using Adam optimizer with learning rate 0.001 for 100 epochs and batch size 32.

For the two more involved datasets (houses and movies), we used conditional feature adding Algorithm \ref{alg:add_feature_conditional} with 10-20 cycles of adding 5 features and removing 2. During the process, when a discovered feature is active for less than 2\% or more than 98\% of the data, it is discarded, resulting in the feature count generally growing until it stabilizes towards the end of the runs. 

Mining prompts are executed by GPT 5, while the extraction prompts are executed with GPT 4.

\subsection{Learning, or rather, recalling binary}
The primary purpose of this experiment was to serve as an explanation of what iteration of the Algorithm \ref{alg:add_feature} tends to do. The semantic items $s \in \{ "0", "1", \ldots ,  "511"\}$, have the corresponding targets $y = \text{str2num}(s)$. We map characteristics $\bz$ to predictions $y$ with a linear function $f(\bz)=\bw^T\bz+b$. 

When we add the first feature by running Algorithm \ref{alg:add_feature}, $q_i$ splits the data into high and low values (above or below the mean). When positive and negative groups are incorporated into the mining prompt (Figure \ref{fig:prompts}) and given to GPT 5, it easily detects the pattern: Depending on the random initialization of $\bw$, the positive group's characteristic is of two sorts: either an equivalent of "numbers 0-255 inclusive", or an equivalent of "numbers greater than 255".

The first feature splits the data into high and low values, and for all items the model will now predict the mean of one of these two subsets: 127.5 for numbers less than 256, and 383.5 for the rest. The next execution of Algorithm \ref{alg:add_feature} will now assign $z_2=1$ and $z_2=0$ to the items based on where the model over- or underestimates $y$. (Due to randomness in initialization, $z_2=1$ could focus on either all of the overshoots or all of the undershoots). So, the positive group will consist of either the upper halves of the $z_1$ groups or the lower halves. For example, the positive group would include all items in \emph{both} range $[0..127]$ and range $[256..383]$. Given these as the positive group, and the rest as negative in the mining prompt, GPT 5 sets the next feature to be the equivalent of \verb|"Integers with the 2^7 bit unset."| As before, due to random initialization, the positive and negative group could be switched, but that would still yield an equivalent feature, as the model simply adjusts weights.

Here is an example of a set of features $\theta_f$ obtained by running Algorithm {\ref{alg:add_feature}} 9 times with $i$ increasing from 1 to 9:
{\small
\begin{itemize}
\setlength\itemsep{0em}
\item Integers greater than or equal to 256 (i.e., with the 2\^{}8 bit set)
\item Integers with the 2\^{}7 bit set (i.e., numbers whose value modulo 256 is between 128 and 255)
\item Integers with the 2\^{}6 (64) bit unset (i.e., numbers whose value modulo 128 is 0--63)
\item Integers with the 2\^{}5 (32) bit set (i.e., numbers whose value modulo 64 is between 32 and 63)
\item Integers with the 2\^{}4 (16) bit unset (i.e., numbers whose value modulo 32 is 0--15)
\item Integers with the 2\^{}3 (8) bit set (i.e., value modulo 16 is between 8 and 15)
\item Integers with the 2\^{}2 (4) bit unset (i.e., numbers whose value modulo 8 is 0--3)
\item Integers with the 2\^{}1 (2) bit set (i.e., numbers congruent to 2 or 3 modulo 4)
\item Integers with the 2\^{}0 (1) bit unset (even numbers)
\end{itemize}
}

And here is another one:
{\small
\begin{itemize}
\setlength\itemsep{0em}
\item Integers from 256 to 511 (inclusive)
\item Integers whose value modulo 256 is in the range 128--255 (i.e., their low byte has its highest bit set)
\item Integers whose value modulo 128 is in the range 0--63 (inclusive)
\item Integers whose value modulo 64 is in the range 0--31 (the lower half of each 64-block)
\item Integers whose value modulo 32 is in the range 16--31 (the upper half of each 32-number block)
\item Integers whose value modulo 16 is in the range 0--7 (the lower half of each 16-number block)
\item Integers whose value modulo 8 is in the range 0--3 (the lower half of each 8-number block)
\item Integers congruent to 2 or 3 modulo 4 (i.e., with the second least significant bit set)
\item Even integers (including 0)
\end{itemize}
}

Both reconstructions are perfect. We see the same for other ground truth function choices of $y$ that are linearly related to the value of the number in the string $s$, $y = a_1 \cdot \text{num2str}(s) + a_2$. The feature discovery is driven by the structure of the algorithm that breaks the data in a coarse-to-fine manner through binary splits, perfectly matching the binary number representation. All that a foundational model has to do is to recognize this in the provided positive/negative splits. In many applications, such coarse-to-fine binary splitting of the data can be problematic, as features may not interact additively. In the next two experiments in more realistic scenarios, we use Algorithm \ref{alg:add_feature_conditional}, which chooses the groups conditional on the assignment of the already discovered features.

\subsection{Hedonic regression of house prices}
\label{sec:houses}
\begin{figure}[h]
\centering
\begin{tikzpicture}[
    imgbox/.style={draw, rounded corners, minimum width=6.9cm, minimum height=6.6cm,
                   align=center, text width=6.5cm},
    jsonbox/.style={draw, rounded corners, minimum width=6.4cm, minimum height=8.0cm,
                    align=left, text width=6.1cm},
    node distance=0.35cm
]

\node[imgbox] (box1) {
  \begin{varwidth}{\linewidth}
  \begin{small}
  \begin{flushleft}

  \textbf{Metadata}\\[-0.5mm]
  {\scriptsize
  Bedrooms: 4; Bathrooms: 4; Area: 4053 sqft; ZIP: 85255; Price: \$869,500
  }\\[1mm]

  \textbf{Listing images}\\[1mm]
  \end{flushleft}

  \begin{center}
    \begin{minipage}{0.97\linewidth}
      \centering
      \setlength{\tabcolsep}{6pt}
      \renewcommand{\arraystretch}{1.10}
      \begin{tabular}{@{}c c@{}}
        \begin{minipage}{0.47\linewidth}
          \centering
          \includegraphics[width=\linewidth]{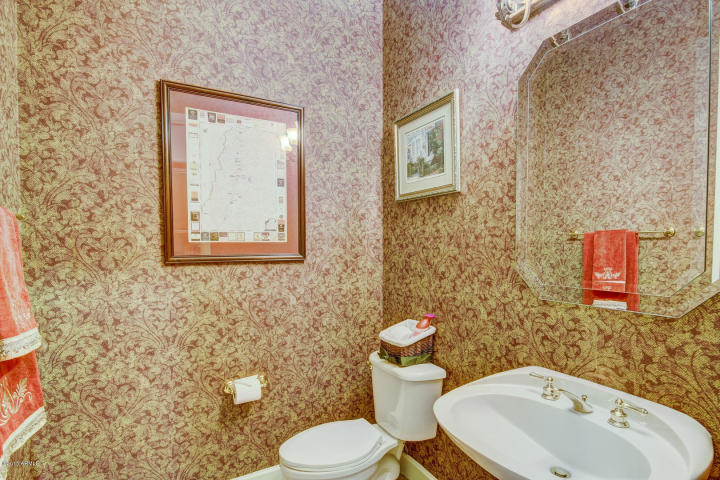}\\[-1mm]
          {\scriptsize\ttfamily bathroom}
        \end{minipage}
        &
        \begin{minipage}{0.47\linewidth}
          \centering
          \includegraphics[width=\linewidth]{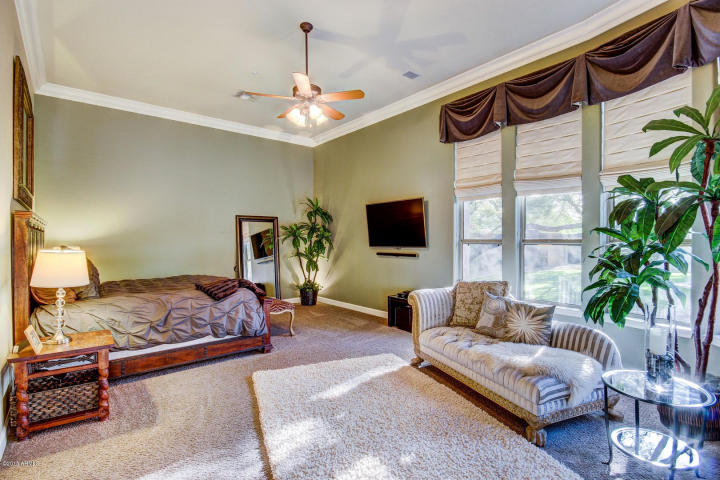}\\[-1mm]
          {\scriptsize\ttfamily bedroom}
        \end{minipage}
        \\[10mm]
        \begin{minipage}{0.47\linewidth}
          \centering
          \includegraphics[width=\linewidth]{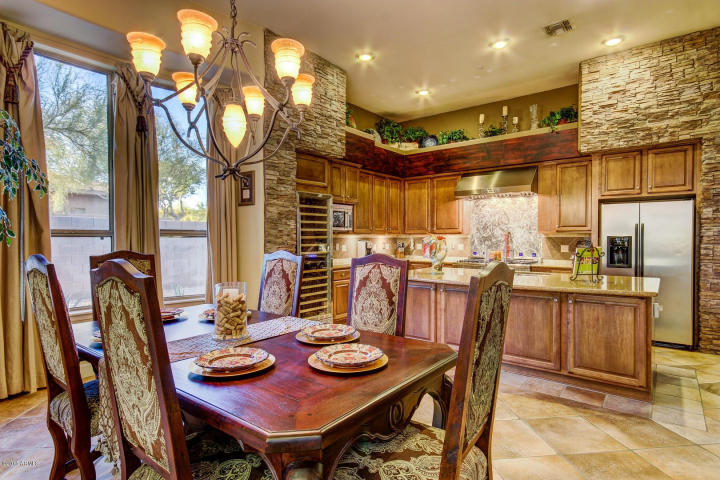}\\[-1mm]
          {\scriptsize\ttfamily kitchen}
        \end{minipage}
        &
        \begin{minipage}{0.47\linewidth}
          \centering
          \includegraphics[width=\linewidth]{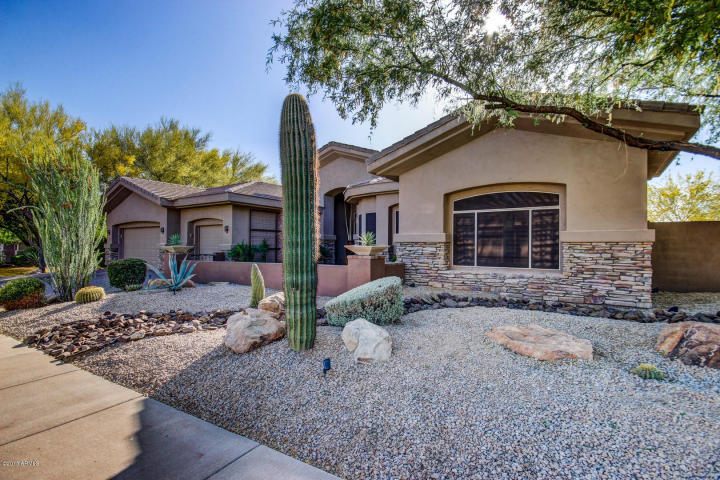}\\[-1mm]
          {\scriptsize\ttfamily frontal}
        \end{minipage}
        \\
      \end{tabular}
    \end{minipage}
  \end{center}

  \end{small}
  \end{varwidth}
};

\node[jsonbox, right=0.35cm of box1] (box2) {
  \begin{varwidth}{\linewidth}
  \begin{small}
  \begin{flushleft}
  \textbf{Semantic item (JSON)}\\[1mm]

  {\ttfamily\scriptsize
  \{\\
  \phantom{aa}"house\_id": 1,\\
  \phantom{aa}"metadata": \{\\
  \phantom{aaaa}"bedrooms": 4,\\
  \phantom{aaaa}"bathrooms": 4.0,\\
  \phantom{aaaa}"area": 4053,\\
  \phantom{aaaa}"zip\_code": "85255",\\
  \phantom{aaaa}"price": \sout{869500}\\
  \phantom{aa}\},\\
  \phantom{aa}"image\_descriptions": \{\\
  \phantom{aaaa}"house\_id": 1,\\
  \phantom{aaaa}"bathroom\_description": "This powder room features ornate, burgundy damask wallpaper...",\\
  \phantom{aaaa}"bedroom\_description": "The bedroom is spacious, with plush carpeting and a soft green accent wall ...",\\
  \phantom{aaaa}"frontal\_description": "The exterior showcases a desert landscape with mature cacti, rocks, and gravel, minimizing maintenance ...",\\
  \phantom{aaaa}"kitchen\_description": "The kitchen and dining area blend rustic and elegant elements. Rich wood cabinetry, stone accents, ..."\\
  \phantom{aa}\}\\
  \}\\
  }

  \end{flushleft}
  \end{small}
  \end{varwidth}
};

\end{tikzpicture}
\caption{House listing images (left) and corresponding MLLM-generated semantic item (right). Strikethrough price indicates that price is withheld from items $s$ in training our model. However, in the alternative to our approach, all items {\bf including prices} were included in baseline prompting, where GPT 5 was given the entire dataset and asked  to generate features predictive of the price.}
\label{fig:house-semantic-item}
\end{figure}

The multi-modal dataset \cite{ahmed2016house} consists of information for 535 houses listed in 2016, the first of which is shown in Figure \ref{fig:house-semantic-item}. While our approach is directly applicable to multi-modal data, to save tokens and avoid limitations on the number of images we can submit in our mining prompt, we first transformed the information to pure text, by running all images through GPT 5 with the prompt \verb|"Describe the listing image in detail."|. Each semantic item is a JSON file with metadata (without the price) and image description, and the target $y$ is the log of the price. (Log price modeling is standard in hedonic regression, as some features multiplicatively affect prices. For example, an upgraded interior is costlier for a larger home, or for a home in an expensive ZIP code.)

\begin{figure}[t]
\centering
\begin{subfigure}[b]{0.497\textwidth}
    \centering
    \includegraphics[width=\textwidth]{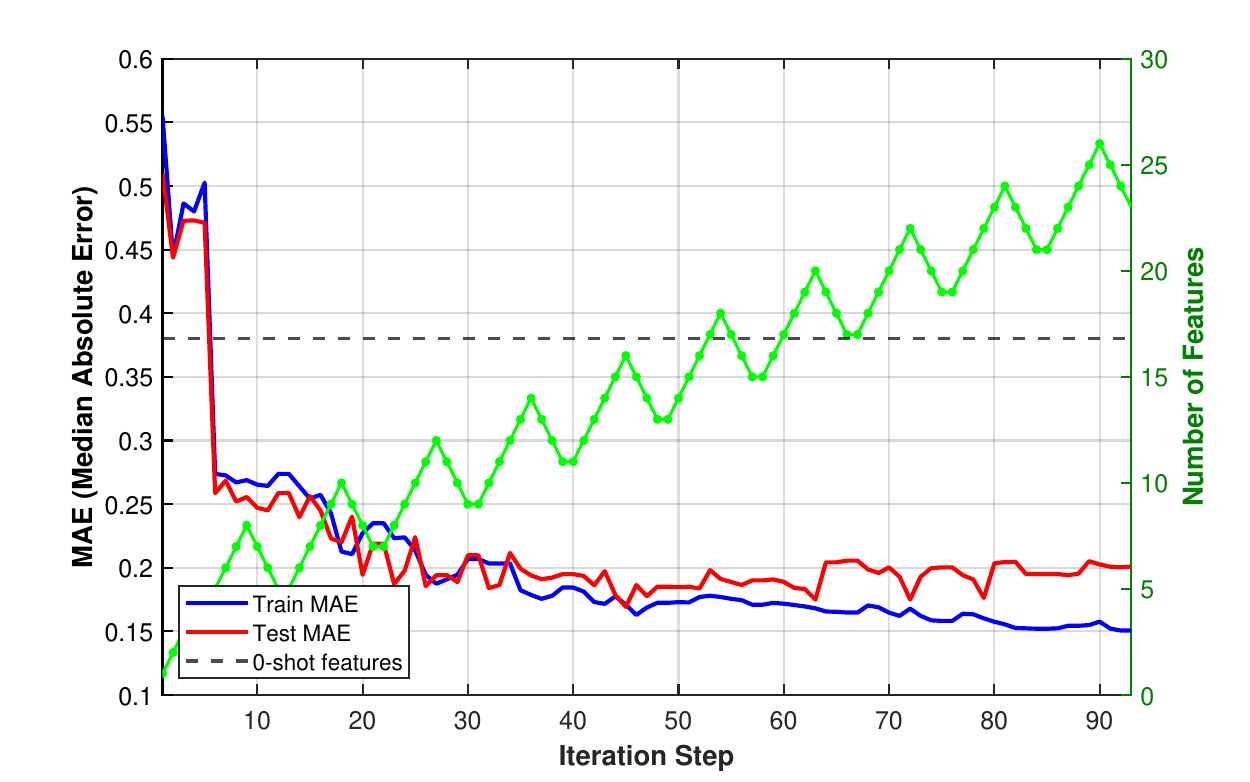}
    \caption{Linear model}
    \label{fig:houses_linear}
\end{subfigure}
\hfill
\begin{subfigure}[b]{0.497\textwidth}
    \centering
    \includegraphics[width=\textwidth]{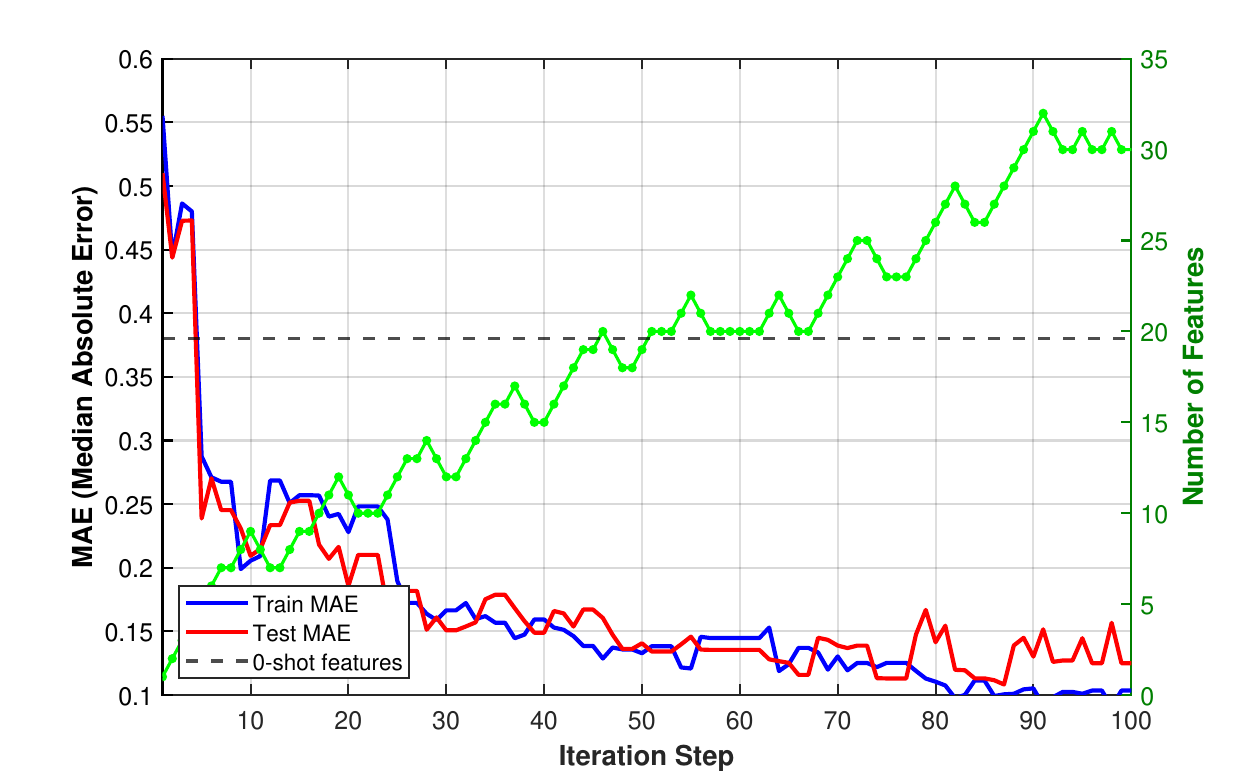}
    \caption{Neural network}
    \label{fig:houses_nn}
\end{subfigure}
\caption{Hedonic regression results on log of house prices. Both plots show the median absolute error (MAE) (left axis) and number of features (right axis) over iterations. The horizontal dashed line shows GPT-5 0-shot baseline (MAE=0.38). The video showing which features are added (blue) or removed in (b) is available at: \url{https://github.com/mjojic/genZ/tree/main/media}.}
\label{fig:houses_results}
\end{figure}

As discussed in Section \ref{sec:experiments}, we fit $y$ using both linear and nonlinear (neural net) forms of the function $f$. In both cases, we supply $f$ with additional real-valued inputs $x_1 = 1, x_2 = \text{area}, x_3 = \text{bedrooms}, x_4 = \text{bathrooms}$, as we do not wish to rediscover these numerical features as we did in the toy experiment in the previous section. However, in the semantic items $s$, full metadata (the number of bedrooms and bathrooms, the area in square feet, but not the price) is still provided, as it can participate in features $\bz$ to account for both nonlinear transformations of price and combined feature effects. For example, in some locales, luxury homes have more bathrooms than bedrooms and that signal can be captured in the discrete $z$ as an indicator of higher priced homes. 

We split the data into the training set (80\% of the data) and the test set. We show the learning curves in Figure \ref{fig:houses_results} for the linear model $f$ and a neural network $f$. The video showing which features are added (blue) or removed (red) in Figure \ref{fig:houses_results} (b) is available at: \url{https://github.com/mjojic/genZ/tree/main/media}. We plot the progress in terms of median absolute error, as this corresponds to how the performance of hedonic regression of house prices tends to be evaluated in industry. For example, Zillow \cite{zillow} reported that its Zestimate for off-market homes has a median error rate of 7.01\%. (Zestimate for on-market homes is much better because it benefits from the known listing price set by an agent; In our experiments we are trying to predict the price without such professional appraisal). As we model log prices, small median absolute errors (MAE) are close to the relative (percentage) error of the actual (not log) price. For example, achieving MAE of 0.10-0.11, as our model does, is equivalent to a median error rate of around 12\%. 

The 0-shot baseline is indicated by the horizontal line. The baseline is computed as follows. First, we provided the \emph{entire} set of semantic items, \emph{including the prices} to GPT 5 and asked for 50 yes/no questions whose answers would be helpful in predicting the price (GPT gave us 52 instead). Then, these are set to be features $\theta_f$, and the pruning Algorithm \ref{alg:prune_feature} is applied repeatedly on the training set using linear $f(\bz,\bx)$, keeping track of the test performance. (Note that the function $f$ still takes the numeric features $x_1, x_2, x_3, x_4$). The horizontal line on the graph represents the best test set performance achieved during pruning. As discussed in Sections \ref{sec:intro}, \ref{sec:experiments}, this baseline is meant to simulate what could be expected from methods that rely on LLM's understanding of the domain, rather than iteratively contrasting items based on modeling error as our approach does. The baseline vastly underperforms, and by comparing the baseline's features with the features our method discovered (Section \ref{sec:house_features}) we can see why. For example, the baseline's features rarely focus on location or the architecture/quality of the build/remodel, and instead capture relatively cosmetic features such as fenced yards or green lawns.

The differences in the learning curves provide another illustration that the hybrid model is adjusting all its parameters in unison during learning. The train and test median absolute error drop together until over-training happens. In the case of the linear model, the training error continues to drop, but the testing error stays stable, while in the case of the nonlinear (neural) model, we see that both train and test errors reach lower levels of around $0.11$. This corresponds to roughly $12\%$ relative error on the raw (not log) price, after which the model begins to overtrain more severely, with training error dropping even lower as the testing error continues to rise. (Note Zillow's 7\% error\cite{zillow}, but based on a different, much larger and richer, proprietary dataset of hand-crafted features).

\subsection{Cold start recommendations: Netflix prize dataset}
\label{sec:netflix}

The cold start problem in collaborative filtering systems refers to the challenge of making predictions for new items or users without sufficient interaction history. We use the classic Netflix prize dataset \cite{bennett2007netflix, netflix-prize-data} to demonstrate how our hybrid model can address the cold start problem for new movies by discovering semantic features that predict collaborative filtering embeddings.

For 480,189 users and 17,770 movies, the dataset provides user-movie ratings (0-5, where 0 indicates that the movie was not rated by that user). We computed 32-dimensional embeddings $\by$ for each movie using singular value decomposition (SVD) of the ratings matrix. It is well understood that such embeddings capture latent user preferences and movie characteristics discovered through collaborative filtering, and are remarkably good at modeling similarity between movies and matching users to movies via simple vector projection of the embeddings. (See also the discussion in Section \ref{sec:association_modeling} on how instead of embeddings the raw matrix could be modeled with \emph{two} different types of semantic items -- users and movies -- using a similar formulation to the one above).

\begin{figure}[h]
\centering
\includegraphics[width=0.7\textwidth]{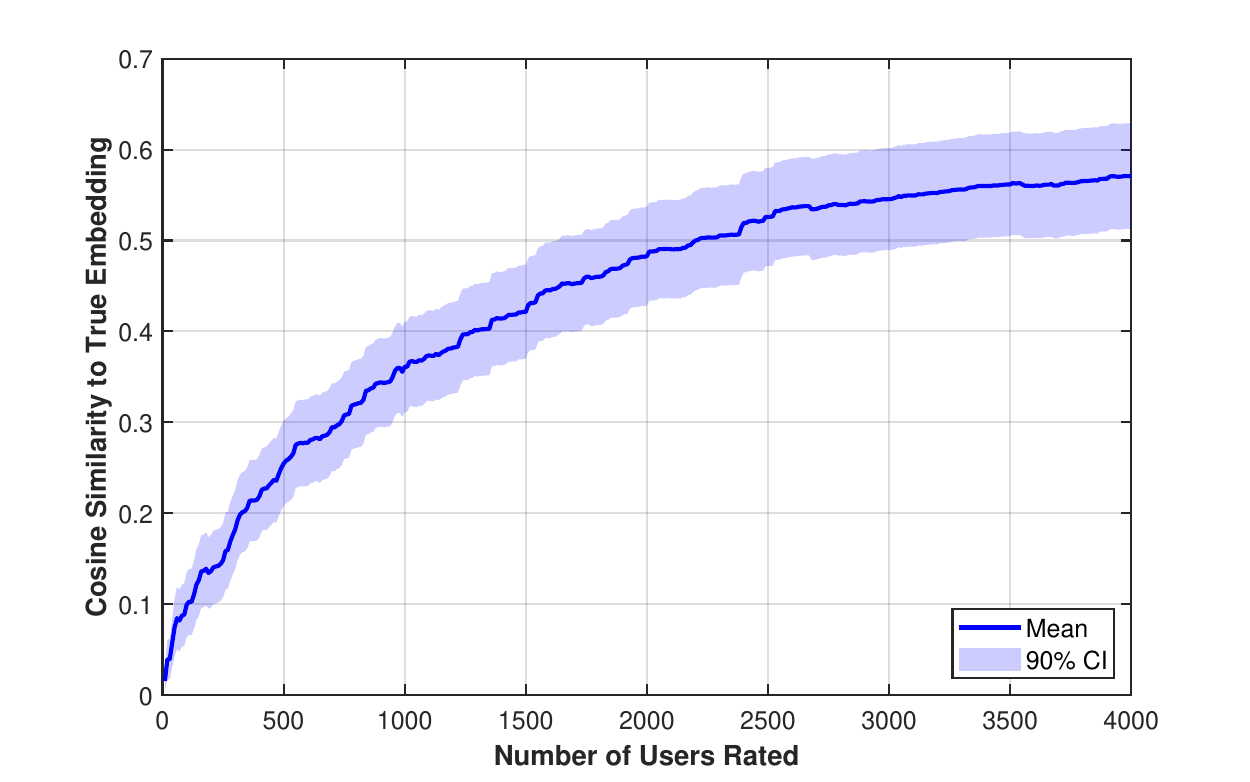}
\caption{Cold start convergence simulation. Cosine similarity between embeddings computed from partial ratings vs. full ratings, as user ratings are gradually added in random order. The shaded region shows 90\% confidence interval over 100 randomly selected movies. Hundreds of ratings are typically needed before embeddings converge, illustrating the severity of the cold start problem that our hybrid model addresses by predicting embeddings from semantic features alone.}
\label{fig:cold_start}
\end{figure}

To illustrate the cold start problem, we simulate adding a new movie with limited exposure to viewers, by removing a single movie and gradually adding user ratings. For a selected movie, we set the values in the ratings matrix to zero for all users, then gradually add users' actual ratings in a random order and recompute the movie's embedding after each addition. Figure \ref{fig:cold_start} shows how the cosine similarity (CS) between the original (full observation) movie embedding and the embedding based on sparse observations improves as we add ratings for up to 4000 users. The plot shows the mean and 90\% confidence interval over 100 randomly selected movies from the dataset. The similarity reaches only about 0.38 after 1000 users and approximately 0.57 after 4000 users. In other words, thousands of users would need to be exposed to the movie, many skipping it, and some rating it, before the CS could grow to around 0.6, leading to more reliable recommendation. We show below that our linear hybrid model can predict embeddings immediately from the semantic items describing the movie, achieving test CS of approximately 0.59 without requiring any user ratings---a performance level that would otherwise require roughly 4000 user ratings to achieve through collaborative filtering alone.

In our experiments, we focused on the 512 most watched movies, again randomly split into training (80\%) and test set. This allowed us to set the semantic items $s$ for these movies to consist only of the movie title and year of release: Modern LLMs are familiar with the most popular movies watched in the period from 1998 to 2006. Otherwise, we would need items $s$ to contain more information, as was the case in the house experiment, or allow the LLM to use a RAG system to retrieve relevant information when executing extraction and mining prompts. The quality of the predictions is measured using cosine similarity (CS) between predicted and true 32-dim embeddings $\by$.

\begin{figure}[h]
\centering
\begin{subfigure}[b]{0.497\textwidth}
    \centering
    \includegraphics[width=\textwidth]{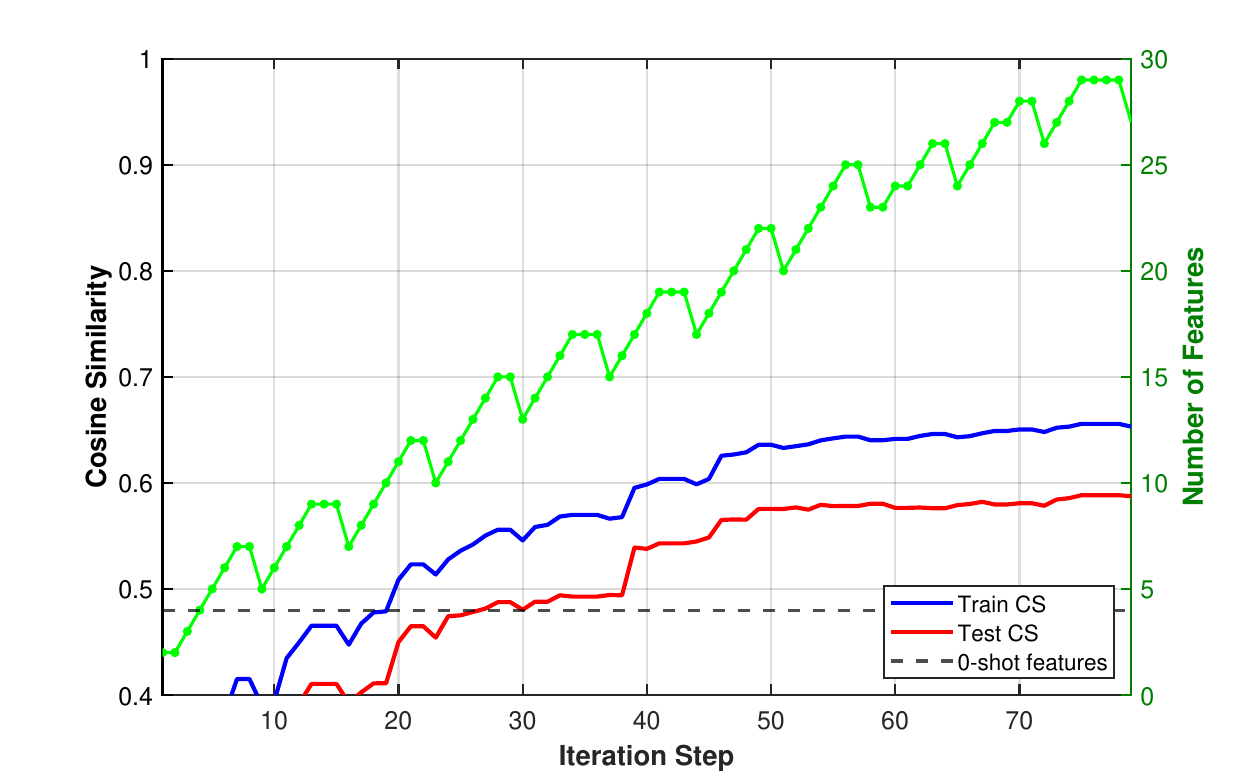}
    \caption{Linear model}
    \label{fig:movies_linear}
\end{subfigure}
\hfill
\begin{subfigure}[b]{0.497\textwidth}
    \centering
    \includegraphics[width=\textwidth]{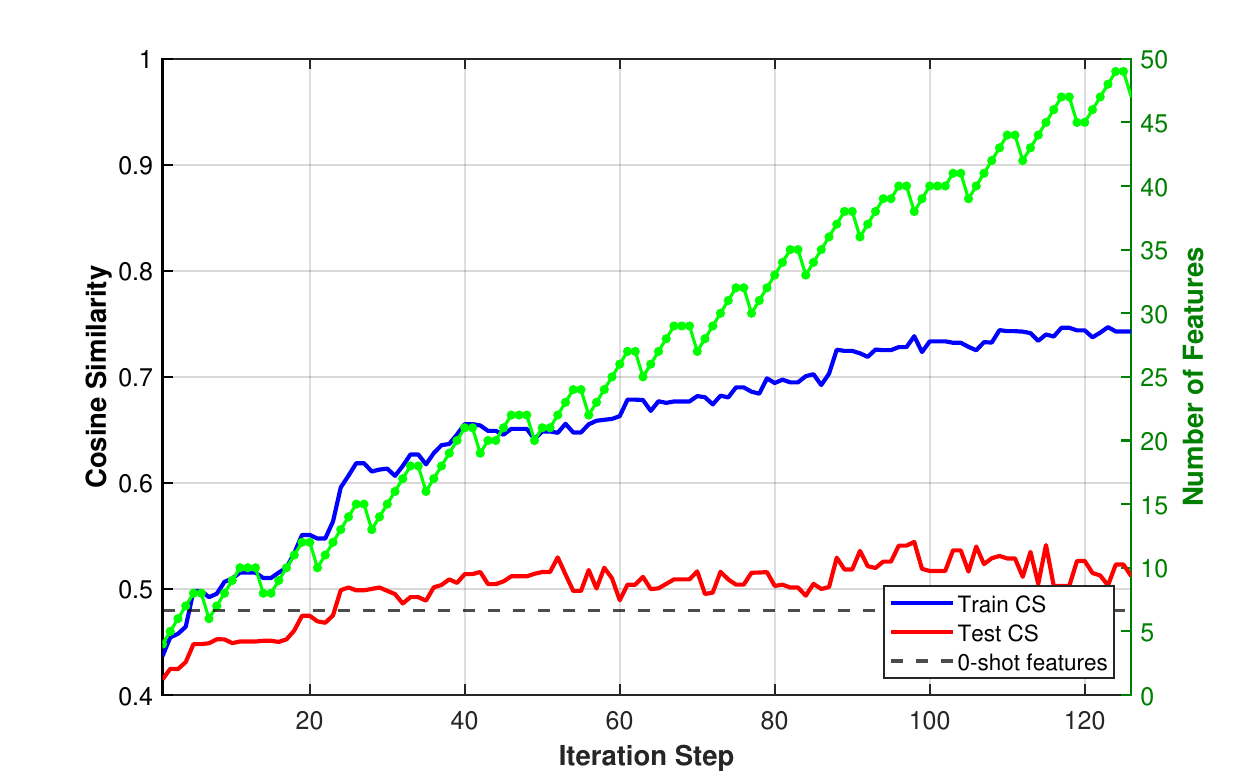}
    \caption{Neural network}
    \label{fig:movies_nn}
\end{subfigure}
\caption{Cold start recommendation results on Netflix dataset. Both plots show cosine similarity (CS) between predicted and true embeddings (left axis) and number of features (right axis) over iterations. The horizontal dashed line shows GPT-5 0-shot baseline (CS=0.5). The linear model achieves better test performance (CS$\approx$0.59) than the neural network (CS$\approx$0.52) despite using fewer features, suggesting that the relationship between semantic features and collaborative filtering embeddings is predominantly linear. The video showing which features are added (blue) or removed in (a) is available at: \url{https://github.com/mjojic/genZ/tree/main/media}.}
\label{fig:movies_results}
\end{figure}

The learning curves are shown in Figure \ref{fig:movies_results} for both linear and neural network models. The video showing which features are added (blue) or removed (red) in Figure \ref{fig:movies_results} (a) is available at: \url{https://github.com/mjojic/genZ/tree/main/media}. As with the houses experiment, we compare against a strong 0-shot baseline (horizontal dashed line at CS=0.48) obtained by asking GPT-5 to suggest 50 features based on the task description and movie titles, then via pruning on the training set find the best feature set in terms of the \emph{test} error, as this is the best the baseline could possibly get for that training-test split. 

The linear model substantially outperforms the 0-shot baseline, reaching a test cosine similarity of approximately 0.59, indicating that the iterative feature discovery process finds movie characteristics more predictive of user preferences than those suggested by the LLM based on task understanding alone. The 0.11 improvement in cosine similarity (from 0.48 to 0.59) is substantial: As the cold start simulation in Figure \ref{fig:cold_start} shows, this difference equals the improvement from adding roughly 2000 additional user ratings. Unlike the house price prediction experiment, using the neural network as the function $f(\bz)$ in the GenZ system leads to more dramatic overfitting, where the test CS score quickly levels off while the train CS continues to grow.

Examining the discovered features reveals interpretable patterns aligned with user preferences, though notably different in character from the 0-shot baseline features (see Section \ref{sec:netflix_features} for complete feature lists). Where the 0-shot features focus on content-based attributes (genre, plot structure, themes, pacing, narrative devices), the discovered features gravitate toward external markers: specific talent (individual actors and composers), multiple fine-grained award distinctions (Best Picture, Best Actor/Actress for specific role, Best Original Screenplay), franchise membership, and surprisingly precise temporal windows (1995-2000, 2004-2005, or post-1970). The John Williams feature is particularly revealing---it cuts across multiple genres to identify a coherent aesthetic or cultural preference that would be invisible to a content-based analysis. This suggests that collaborative filtering embeddings capture user similarity not through "stories like this" but through shared preferences for prestige signals, specific creative talent, and cultural/temporal cohorts (at least in this dataset). In another dataset, focused on a different selection of movies, or a different time frame, or a different cross-section of users, the system would likely discover different emergent structure in collective viewing, likely somewhat in disagreement with the conventional theories of content similarity.

\section{Conclusions}
The differences and similarities of the learning curves relative to the baseline in both experiments indicate that components of the model jointly adjust to different semantics of the data as well as target statistics and their relationships to semantics. Even though the models used here (GPT 4/5) are familiar with the semantics of the data, their ability to predict the real-valued, possibly multi-dimensional targets benefits from the rest of the system and the modeling choices. For example, while a neural network better models interactions of features needed to predict house prices, the linear model is better suited for modeling movie embeddings. In both applications, the learning curves show classic behavior of iterative improvement until over-training. On the other hand, features generated by the LLMs alone are highly redundant preventing both effective prediction and overtraining. As evident in the discovered features (Section \ref{sec:features}), our system focuses on properties of the semantic items that fit the statistics of the links $s\rightarrow\by$, rather than 'conventional wisdom' of a zero-shot approach: For houses, the system discovers that location and quality of build/remodel matters most in the data it was given for listings in 2016, while for movies, the system discovers the features important to the audience in the period 1998-2006, such as the difference between newer and older content, importance of franchises, specific actors/directors, and so on.  

\bibliographystyle{iclr2026_conference}
\bibliography{genaiz}

\newpage
\appendix

\section{Learned features}
\label{sec:features}
\subsection{Learned features for the house dataset}
\label{sec:house_features}

The following are the features discovered by GenZ with the nonlinear (neural net) $f(\bz)$ (on top of autoamtically supplied numerical values $x_1, x_2,x_3, x_4$ for intercept, area, number of bedrooms and number of bathrooms), reaching the test and train median relative error around 12\% of the raw target price (Fig. \ref{fig:houses_nn}:

{\small
\begin{itemize}
\setlength\itemsep{0em}
\item Site-built construction (not a manufactured/mobile home)
\item Located in Arizona desert communities (Scottsdale/Carefree area: zip codes 85255, 85262, 85377)
\item Located in the 92880 zip code (Eastvale/Corona area)
\item Located in the 62234 zip code (Collinsville, IL area)
\item Located in the 94531 zip code (Antioch, CA area)
\item Located in the 92276 zip code (Desert Hot Springs, CA)
\item Located in the 93111 zip code (Goleta/Santa Barbara area)
\item Located in the 81524 zip code (Loma, CO area)
\item Located in the 93510 zip code (Acton, CA area)
\item Located in the 91901 zip code (Alpine, CA area)
\item Bathrooms feature retro/vintage elements (e.g., colored fixtures or tile like powder blue/pink, clawfoot tubs, or \item checkered floors)
\item Located in the 92677 zip code (Laguna Niguel, CA)
\item Bathrooms featuring Hollywood-style bulb strip lighting above the mirror
\item Located in the 94501 zip code (Alameda, CA)
\item Homes featuring plantation shutters on windows
\item Located in the 96019 zip code
\item Located in Chicago-area suburbs with zip codes starting with 600 (e.g., 60002, 60016, 60046)
\item Bathrooms feature built-in vanities with cabinetry (no pedestal sinks)
\item Interior walls feature wood paneling
\item Kitchen does not include an island (no central island present)
\item Home includes a carport for covered parking (rather than an attached garage)
\item Located in zip codes starting with 90 (Los Angeles/Long Beach area)
\item Home includes a three-car garage
\end{itemize}
}

In Fig. \ref{fig:houses_nn}, the horizontal line represents the minimum test error after training on the training set with ever-reducing set of baseline features obtained in what we refer to as 0-shot manner. Specifically, to create those baseline features, GPT 5 was given the entire dataset (training and test) together {\bf with prices} and asked to hypothesize features useful for price prediction. The following are those 52 features from the 0-shot baseline used in feature selection:
{\small
\begin{itemize}
\setlength\itemsep{0em}
\item House has 4 or more bedrooms
\item House has 5 or more bedrooms
\item House has 3 or more full bathrooms
\item House has 4 or more full bathrooms
\item House has an attached garage
\item House has a 3-car (or larger) garage
\item House has a circular driveway
\item House has a gated entry (wrought-iron gate/entry gate)
\item House has a covered front porch
\item House has a covered patio
\item House has a pergola for shade
\item House has a raised deck
\item House has a balcony
\item House has a private swimming pool
\item House has a resort-style pool with water features
\item House has waterfront access (dock/marina/pond view)
\item House has mountain or panoramic scenic views
\item House has extensive desert/xeriscape landscaping (gravel/cacti/rocks)
\item House has a green lawn (grass yard)
\item House has a fenced yard
\item House has solar panels on the roof
\item House has a tile roof (e.g., clay/Spanish tile)
\item House has stucco exterior walls
\item House has exterior stone/stacked-stone accents
\item House has Mediterranean-inspired architecture (arches/turret/columns)
\item House has Craftsman-style exterior elements (stone columns/shingle siding)
\item House has an open-concept kitchen/living layout
\item House has a large kitchen island
\item House has a breakfast bar or peninsula seating in the kitchen
\item House has granite or stone kitchen countertops
\item House has a tile or mosaic kitchen backsplash
\item House has stainless steel kitchen appliances
\item House has a double oven
\item House has recessed lighting
\item House has decorative chandelier lighting
\item House has hardwood flooring in main living areas
\item House has tile flooring in main living areas
\item House has wall-to-wall carpeting in bedrooms
\item House has plantation shutters
\item House has vaulted or tray ceilings
\item House has crown molding
\item House has exposed wood ceiling beams
\item House has a fireplace (in any room)
\item House has a fireplace in the primary bedroom
\item House has a walk-in shower with glass enclosure
\item House has a soaking tub in the primary bathroom
\item House has a jetted tub (spa tub)
\item House has a freestanding bathtub
\item House has a double vanity (two sinks) in a bathroom
\item House has glass block windows in a bathroom
\item House has a skylight
\item House has a built-in desk or built-in shelving/storage cabinetry
\end{itemize}
}

\subsection{Discovered features for Netflix dataset}
\label{sec:netflix_features}

The following 27 features were discovered by our approach with the linear model $f(\bz)$, reaching test CS of 0.59 (Fig. \ref{fig:movies_linear}):
{\small
\begin{itemize}
\setlength\itemsep{0em}
\item Not science fiction (no movies centered on aliens, space, or futuristic tech)
\item Not released between 1995 and 2000 (inclusive)
\item Not primarily comedies
\item Not primarily romance-focused films (i.e., avoids romantic dramas and romantic comedies)
\item Mostly PG-13/PG-rated mainstream releases (i.e., generally avoids hard-R films)
\item Not animated (live-action films)
\item Not primarily action-adventure, heist, or spy thrillers (i.e., avoids high-octane studio action crowd-pleasers)
\item Not horror (i.e., avoids horror and horror-leaning supernatural thrillers)
\item Includes many Academy Award Best Picture nominees/winners (i.e., more prestige/canon titles)
\item Includes a notable number of pre-1995 titles (1960s–early ’90s), whereas the negatives are almost entirely 1995–2004 releases
\item Primarily entries in major film franchises or multi‑film series (including first installments that launched franchises)
\item Includes multiple war films set amid real historical conflicts (e.g., WWII/Vietnam), which are absent in the negative group
\item Not crude/raunchy, man‑child/SNL‑style comedies (when comedic, entries skew quirky/indie or gentle rather than broad/frat humor)
\item Skews toward female-led or women-centered narratives (e.g., Legally Blonde, The First Wives Club, Freaky Friday, The Pelican Brief, Lara Croft), whereas the negative group is predominantly male-led
\item Excludes 2004–2005 releases (the positives stop at 2003, while many negatives are from 2004–2005)
\item Includes multiple films centered on high school/college life or set primarily in educational settings (e.g., The Breakfast Club, Sixteen Candles, Fast Times at Ridgemont High, Billy Madison, Good Will Hunting, Top Gun), a theme largely absent from the negatives
\item Not children/family-oriented fare (avoids Disney-style live‑action family movies and musicals; skews toward teen/adult titles instead)
\item Not quirky/indie/cult, auteur‑driven or formally experimental films; the positives skew toward conventional, mainstream studio storytelling
\item Not ancient/medieval ‘sword‑and‑sandal’ or knightly epics (i.e., avoids Greco‑Roman/medieval battlefield sagas like Braveheart or Troy)
\item Includes multiple films scored by John Williams (e.g., Return of the Jedi, Raiders of the Lost Ark, E.T., Harry Potter and the Sorcerer’s Stone), whereas none of the negatives are Williams-scored
\item Includes multiple films that earned their lead performer an Academy Award (Best Actor/Best Actress) for that specific role (e.g., Monster, Training Day, Monster's Ball, Dead Man Walking), whereas the negatives have few or none
\item Standard theatrical releases only (no titles labeled as Special/Extended/Anniversary/Platinum Edition)
\item Includes multiple films starring Angelina Jolie or Brad Pitt (none of the negatives star either)
\item Includes multiple films that won or were nominated for the Academy Award for Best Original Screenplay
\item No pre-1970 releases (all positives are 1972 or later, while the negatives include several 1950s–1960s films)
\item Includes multiple films starring Tom Hanks (e.g., Forrest Gump, Sleepless in Seattle), whereas none of the negatives feature him
\end{itemize}
}

On the other hand, a 0-shot approach produces very different features. The following 50 features were suggested by GPT-5 based on task description, then pruned to achieve the baseline test CS of 0.48:

{\small
\begin{itemize}
\setlength\itemsep{0em}
\item Is animated
\item Is part of a franchise/sequel/prequel/remake
\item Is based on a true story or real events
\item Is adapted from a non-comic book or short story
\item Is adapted from a comic book/graphic novel
\item Features a superhero or superpowered protagonist
\item Is set primarily in a historical period (pre-1950)
\item Is set primarily in the future
\item Is set primarily in a contemporary/modern time
\item Is set primarily outside the United States
\item Has predominantly non-English dialogue
\item Has an ensemble cast with multiple co-leads
\item Has a female lead or co-lead driving the story
\item Centers on a male buddy duo
\item Has a strong, central romantic plot
\item Has a dominant comedic tone
\item Is action-driven with frequent set pieces
\item Contains graphic/intense violence
\item Contains significant gunplay
\item Contains extensive swordplay or martial arts
\item Features large-scale war or battle scenes
\item Contains high-speed car chases/vehicular action
\item Centers on a heist/caper
\item Centers on spy/espionage activities
\item Centers on a crime investigation/detective story
\item Centers on courtroom/legal drama
\item Centers on supernatural or paranormal elements
\item Centers on science fiction elements (tech, AI, space)
\item Centers on fantasy worldbuilding (magic, mythical beings)
\item Has prominent horror elements and scares
\item Hinges on a mystery with a major twist/reveal
\item Uses a nonlinear or fractured timeline
\item Uses an unreliable narrator or ambiguous reality
\item Has slow-burn, mood-driven pacing
\item Has fast-paced, high-energy pacing
\item Has a feel-good/optimistic tone with an uplifting resolution
\item Has a tragic or bittersweet ending
\item Has a dark or cynical tone
\item Explores strong themes of redemption or forgiveness
\item Is driven by a revenge motivation
\item Is a coming-of-age or personal growth story
\item Features a prominent mentor–protégé relationship
\item Centers on friendship/buddy dynamics
\item Centers on family relationships (parents/children/siblings)
\item Is a musical (characters sing to advance the story)
\item Is sports-centered
\item Has a military setting or procedural focus
\item Relies heavily on CGI/visual effects spectacle
\item Won at least one Academy Award
\item Has a runtime over 140 minutes
\end{itemize}
\section{Tractably summing out z in joint updates for the linear observation model (factor analysis)}
\label{sec:linear_observation_model}
In our experiments, we used a generic version of the GenZ learning algorithm with the functional link $\bz\rightarrow\by$ arbitrary (plug-and-play). Instead of computing exact expectations of form $E_{q(\bz)} g(\bz)$, we just evaluated $g$ at the mode of $q$. While this allowed us to run the same algorithm for the linear and nonlinear (neural net) models, we noted that in some cases these expectations can be computed tractably under the full distribution $q$. Linear model is such a case, and we derive it here for illustration.  

A linear link between $\bz$ and $\by$ takes a form
\begin{equation}
    p(\by|\bz, \theta_y=\{\bLambda,\sigma^2\})={\cal N}|(\by| \bLambda\bz, \sigma^2I),
    \label{eq:linear_obs}
\end{equation}
where $\bz$ is now $(n_f+1)\times 1$ feature vector with an additional element which is always $1$, $\by$ is a $n_d\times 1$ observation vector, and $\bLambda$ is an $n_d \times (n_f+1)$ parameter matrix. (Note that the overall model is still nonlinear because the relationships among features $\bz$ in $p(\bz|s)$ are nonlinear). The additional constant element $z_{n_f+1}=1$ in $\bz$ allows for learning the bias vector within $\bLambda$. 
\begin{eqnarray}
    T_1=E_q[\log p(\by|\bz)]&=&E_q\big[-\frac{1}{2\sigma^2}(y-\bLambda\bz)^T(y-\bLambda\bz)\big]-\frac{n_d}{2}\log 2\pi\sigma^2 \nonumber\\
    &=&-\frac{1}{2\sigma^2}E_q\big[\by^T\by-2\by^T\bLambda \bz +\bz^T\bLambda^T\bLambda\bz\big] -\frac{n_d}{2}\log 2\pi\sigma^2 \nonumber\\
    &=& -\frac{1}{2\sigma^2}\big(\by^T\by-2\by^T\bLambda E_q[\bz] +E_q[tr(\bLambda\bz\bz^T\bLambda^T]\big) -\frac{n_d}{2}\log 2\pi\sigma^2 \nonumber\\
    &=& -\frac{1}{2\sigma^2}\big(\by^T\by-2\by^T\bLambda E_q[\bz] +tr(\bLambda E_q[\bz\bz^T]\bLambda^T)\big) -\frac{n_d}{2}\log 2\pi\sigma^2
\end{eqnarray}

We use the approximate posterior limited to the form $q(\bz)=\prod_i q(z_i)$. Defining $q(z_i=1)=q_i$, $q(z_i=0)=1-q_i$, and $\bq=[q_1, q_2,... q_{n_f}, 1]^T$ (as $z_{n_f+1} := 1$),  we have
\begin{eqnarray}
    E_q[\bz]&=&\bq, \\
    E_q[\bz\bz^T]&=& \bq\bq^T+\diag(\bq-\bq \odot \bq),
    \label{eq:expectations}
\end{eqnarray}
where $\odot$ denotes point-wise multiplication. (To see this, note that for nondiagonal elements $z_i z_j$ in $\bz\bz^T$ we have $E[z_i z_j]=q_i q_j$ but the expectation for the diagonal elements is $E[z_i^2]=q_i$, rather than $q_i^2$). Thus,
\begin{eqnarray}
    tr(\bLambda E_q[\bz\bz^T]\bLambda^T)&=&\tr(\bLambda \bq\bq^T \bLambda^T)+\tr(\bLambda\diag(\bq-\bq \odot \bq)\bLambda^T)\big) \nonumber\\
    &=&\bq^T \bLambda^T\bLambda \bq +\tr(\bLambda\diag(\bq-\bq \odot \bq)\bLambda^T)\big),
\end{eqnarray}
\begin{equation}
    T_1=-\frac{1}{2\sigma^2}(\by-\bLambda\bq)^T(\by-\bLambda\bq)-\frac{1}{2\sigma^2}\tr(\bLambda\diag(\bq-\bq \odot \bq)\bLambda^T)\big)-\frac{n_d}{2}\log 2\pi\sigma^2.
    \label{eq:T1}
\end{equation}
This is the only term that involves the parameter $\theta_y=\bLambda$, and in the M step, for a given $\bq$ we will solve for $\bLambda$ by maximizing it. The derivation is also useful for the E step, in which we estimate $\bq$. 

With our choice of factorized posterior, the feature term $T_2$ simplifies to 
\begin{eqnarray}
    T_2=E_q[\log p(\bz|s)]=\sum_i \sum_{z_i}q(z_i)\log p(z_i|s)=\sum_i q_i\ell^1_i+(1-q_i)\ell^0_i,
    \label{eq:T2a}
\end{eqnarray}
with $\ell^1_i$ and $\ell^0_i$ defined in (\ref{eq:feature_model}). Finally, the entropy part of the bound  becomes
\begin{equation}
    T_3=-E_q[\log q(\bz)]=-\sum_i q_i\log q_i + (1-q_i)\log(1-q_i).
    \label{eq:T3a}
\end{equation}
To perform inference, i.e. to approximate the posterior $p(\bz|\by,s)$, we optimize the bound $B=T1+T2+T3$ wrt to $\{q_i\}$, the posterior probabilities $q(z_i=1)$. This is a part of the (variational) E-step needed to compute the expectations (\ref{eq:expectations}) used in the observation part $T_1$ in (\ref{eq:T1}). By maximizing $T_1$, and therefore the log likelihood bound (\ref{eq:bound}), wrt to $\bLambda$ ans $\sigma^2$ we then perform a part of the M step related to the parameters of the observation conditional (\ref{eq:linear_obs}). In the M-step we can also maximize the bound wrt uncertainty parameters $\theta_e=\{p^e_i\}$ of the feature model (\ref{eq:feature_model}), and we use the mining prompt in Fig. \ref{fig:prompt_mining} to re-estimate the semantic model parameters, the feature descriptions $\theta_f$ as discussed in Sect. \ref{sec:prompt_mining}. While the optimization can be done with off-the-shelf optimization packages, we derive EM updates here.

{\bf E step:}
For a given $(s, \by)$ pair we optimize $q_i$ one at a time, keeping the rest fixed. A simple way to approximate $q_i$ for a tight bound is to introduce vectors $\bq^{+i}$, and $\bq^{-i}$ which are both equal to $\bq$ in all entries but the $i$-th. In the i-th entry, $\bq^{+i}$ equals 1, and $\bq^{-i}$ equals 0. In other words, the two are parameter vectors for posterior distributions where the i-th feature $z_i$ is forced to be either 1 or 0 with certainty.
Given the above derivations, we can thus compute the approximate log likelihood
\begin{eqnarray}
    \log p(\by|z_i=1,s)&=&\log\sum_{\{z_j\}_j \neq_i}p(\by|\{z_j\}_j, z_i=1, s) \nonumber\\
    &\approx&E_{\bq^{+i}}[\log p(\by|\bz)] +E_{\bq^{+i}}[\log p(\bz|s)] - E_{\bq^{+i}}[\log q(\bz)],
\end{eqnarray}
and similarly for $p(\by|z_i=0,s)$, but using expectations wrt $\bq^{-i}$. In each case the three terms are computed as above in (\ref{eq:T1}), (\ref{eq:T2a}), and (\ref{eq:T3a}), but with the appropriate distribution parameter vector $\bq^{+i}$ or $\bq^{-i}$ replacing $\bq$. The posterior parameter $q_i=\frac{p(\by|z_i=1,s)}{p(\by|z_i=1,s)+p(\by|z_i=0,s)}$ is thus approximated as
\begin{equation}
    q_i=\frac{e^{E_{\bq^{+i}}[\log p(\by|\bz)] +E_{\bq^{+i}}[\log p(\bz|s)] - E_{\bq^{+i}}[\log q(\bz)]}}{e^{E_{\bq^{+i}}[\log p(\by|\bz)] +E_{\bq^{+i}}[\log p(\bz|s)] - E_{\bq^{+i}}[\log q(\bz)]}+e^{E_{\bq^{-i}}[\log p(\by|\bz)] +E_{\bq^{-i}}[\log p(\bz|s)] - E_{\bq^{-i}}[\log q(\bz)]}}
\end{equation}
The entropy terms for $\bq^{+i}$ and $\bq^{-i}$ are the same: Applying equation (\ref{eq:T3}) to $\bq^{+i}$ and $\bq^{-i}$ we see that all summands are the same as $\bq^{+i}_j = \bq^{-i}_j$, except for the $i$-th summand which is zero in both cases as $\bq^{+i}_i=1$ and $\bq^{-i}_i=0$. Therefore the entropy terms divide out. The feature terms $E_{\bq^{+i}}[\log p(\bz|s)]$ $E_{\bq^{-i}}[\log p(\bz|s)]$ are not the same. The summands are still the same for $i\neq j$, but the $i$-th terms are different: They are $\ell^1_i$ and $\ell^0_i$, respectively. Therefore, we can divide out the shared parts. The expression simplifies to
\begin{equation}
    q_i=\frac{e^{E_{\bq^{+i}}[\log p(\by|\bz)] +\ell^1_i]}}{e^{E_{\bq^{+i}}[\log p(\by|\bz)] +\ell^1_i}+e^{E_{\bq^{-i}}[\log p(\by|\bz)]+\ell^i_0}}=\frac{1}{1+e^{E_{\bq^{-i}}[\log p(\by|\bz)]-E_{\bq^{+i}}[\log p(\by|\bz)]+\ell^0_i-\ell^1_i}}
\end{equation}
To get all parameters $q_i$ these equations need to be iterated, but the iterations can be interspersed with parameter updates of the M step. In this update rule we see that the foundational model's prediction of $z_i$ is balanced with the observation $\by$ which may be better explained if the different $z_i$ is chosen. As discussed in Sect. \ref{sec:prompt_mining}, this is a signal needed to update the feature descriptor ${\theta_f}_i$. For example, if $\bf y$ is a movie embedding vector, then the presence or absence of the feature "historical war movie" helps explain explain, through $\Lambda$, the variation in $\by$, which in turn reflect user preferences. If the users do not in fact have preference specifically for historical war movies, but instead for movies based on real historical events, war or not, then this feature will only partially explain the variation, and the posterior might indicate the positive feature, $z_i=1$, with high probability for a movie which is about historical events but the foundational model correctly classified as not a "historical war movie."

{\bf M step:}
While the E-step is performed independently for each item $s$, in the M step we consider a collection of item-onservation pairs $(s^t,\by^t)$, and the posterior distributions $\bq^t$ from the E-step, and optimize the sum of the bounds $B^t$. From (\ref{eq:T1}), we see that
\begin{equation}
    \frac{\partial \big(\sum_t B^t\big)}{\partial \bLambda} =-\frac{1}{\sigma^2}\sum_t\bigg((\by^t-\bLambda \bq^t)(\bq^t)^T+\bLambda\diag(\bq^t-\bq^t\odot \bq^t)\bigg)
\end{equation}
Setting the derivative to zero we derive the update
\begin{equation}
    \bLambda=\bigg(\sum_t \by^t(\bq^t)^T\bigg)\bigg(\sum_t \bq^t(\bq^t)^T+\diag(\bq^t-\bq^t\odot \bq^t)\bigg)^{-1}
\end{equation}
The noise parameter $\sigma^2$ is optimized when
\begin{equation}
    \sigma^2=\frac{1}{n_d T} \sum_{t=1}^T (\by^t-\bLambda\bq^t)^T(\by-\bLambda\bq^t)+\tr(\bLambda\diag(\bq^t-\bq^t \odot \bq^t)\bLambda^T).
\end{equation}
We can (and in our experiments do) optimize the feature uncertainty parameters $p^e_i$. These parameters could in principle be extracted from the foundational model using token probabilities or its own claim of confidence in the response (see prompt in Fig.\ref{fig:prompt_mining}). However, the foundational model's own uncertainty only captures its confidence about judging the characteristic. In our hybrid model, we can have $p_e^i$ model additional uncertainty that the feature is well-defined. This is especially useful during learning as features evolve. As we discussed above using the "historical war movie" feature as an example, in the E step the posterior may diverge from the foundational model's feature classification if the feature only approximately aligns with true variation in observations. To estimate, and then reuse, the full uncertainty, we can maximize the bound wrt $p^e_i$, to arrive at:
\begin{equation}
    p^e_i =\frac{1}{T}\sum_{t=1}^T q_i^t[h_i^t=0]+(1-q_i^t)[h_i^t=1],
\end{equation}
where $h_i^t$ are the binary responses of the foundational model for feature i in the semantic item $s^t$. (To derive, see (\ref{eq:feature_model}), although the update is intuitive, as it simply sums the evidence of error as indicated by the posterior).

Finally, to update the feature descriptors ${\theta_f}_i$, we sample or threshold the posteriors $q^t_i$ to get some positive $z_i^t=1$, and negative $z_i^t=0$ examples.  We then use exemplars from each group to form the feature mining prompt in Fig. \ref{fig:prompt_mining}. The fondational model's response in the "characteristic" field is the updated feature description ${\theta_f}_i$.

\section{Statistical modeling of co-occurrence/association/preferences between two sets of items}
\label{sec:association_modeling}

 We studied a model that links a semantic description of an item $s$ with an observation vector $\by$. Among other applications, the approach allows us to find features that determine user preferences for movies \emph{if $\by$ are movie embeddings inferred from movie-user observation matrix}. Here we show that such an approach is a simplification of a model that involves two sets of interacting items (e.g. movies/viewers, products/buyers, plants/geographic areas, images/captions, congress members/bills). That model (without the simplification) can also be learned from data.

Suppose that the matrix of the pairwise interactions $x_{t_1,t_2}$ is observed for item pairs from the two groups $(s_1^{t_1},s_2^{t_2})$. We would then model the joint distribution $p(x|\bz_1,\bz_2)p(\bz_1|s_1) p(\bz_2|s_2)$ where, for example, $s_1$ would be a textual description (or an image!) of a plant and $s_2$ a textual description (or a satellite observation!) of a geographic area, and the scalar $x$ would be the presence (or thriving) signal for the combination. We could develop a number of different complex models with suitable latent variables (in addition to $\bz$), but honoring the traditionally very successful collaborative filtering methods for dealing with such data, we will focus on embedding methods (which typically rely on matrix decomposition methods, at least in recommender systems).

Thus, for each set of items our latent variables $\bz$ should determine a real-valued embedding $\by$ probabilistically, through a distribution $p(\by|\bz)$.

Then, given the low-dimensional embeddings $\by_1, \by_2$ for the items in the pair (e.g. a movie and a user), we would model the association variable as $\by \approx \bu_1^T\bu_2$:
\begin{equation}
    \log p(x|\by_1,\by_2)= \log {\cal N}(x; \by_1^T\by_2, \sigma^2_x)=-\frac{1}{2\sigma_y^2}(x-\by_1^T\by_2)^2-\frac{1}{2}\log(2\pi\sigma^2_x)
    \label{eq:pyzz}
\end{equation}

In the model over observed association strength $x$, item embeddings $\by_1$ and $\by_2$, each based on their own set's feature vectors $\bz_1$, $\bz_2$, which themselves are probabilistically dependent on the items semantics $s_1, s_2$ the likelihood can be approximated as
\begin{eqnarray}
 \log p(x|s_1, s_2)&=&\log \sum_{\bz_1, \bz_2, \by_1, \by_2} p(x|\by_1,\by_2)p(\by_1|\bz_1)p(\by_2|\bz_2)p(\bz_1|s_1) p(\bz_2|s_2) \nonumber\\
 &\geq& E_{q(\bz_1, \bz_2, \by_1,\by_2)}\bigg[\log \frac{p(x|\by_1,\by_2)p(\by_1|\bz_1)p(\by_2|\bz_2)p(\bz_1|s_1) p(\bz_2|s_2)}{q(\bz_1, \bz_2,\by_1,\by_2)}\bigg]
\end{eqnarray}
where $q(\bz_1,\bz_2,\by_1,\by_2)$ is an approximate posterior. If we assume a factored posterior $q=q(\bz_1)q(\bz_2)q(\by_1)q(\by_2)$, and that the posterior over embeddings Dirac $q(\by_1)=\delta(\by_1-\hat{\by}_1)$, $q(\by_2)=\delta(\by_2-\hat{\by}_2)$ then based on (\ref{eq:pyzz}):

\begin{math}
    \log p(x|s_1, s_2)\approx -\frac{1}{2\sigma_x^2}(x-\hat{\by}_1^T\hat{\by}_2)^2-\frac{1}{2}\log(2\pi\sigma^2_y)+ E_q\big[\log \frac{p(\by_1,\bz_1|s_1)}{q(\bz_1)}\big]+E_q \big[\frac{\log p(\by_2,\bz_2|s_2)}{q(\bz_2)}\big],
\end{math}

where the expected embeddings are $\hat{\bu_1}=E_{q(\bz_1)}[\bu_1]$, $\hat{\bu_2}=E_{q(\bz_2)}[\bu_2]$
If we observe the full association matrix Y, e.g. over $n_1\approx18000$ movies, indexed by $i_1$ and $n_2\approx 500000$ users indexed by $i_2$, then the approximate log likelihood of the data, broken into two terms is:

\begin{eqnarray}
    L&=& -\frac{1}{2\sigma_x^2}\sum_{t_1, t_2}(x_{t_1,t_2}-(\hat{\by}_1^{t_1})^T\hat{\by}_2^{t_2})^2-\frac{n_1 n_2}{2}\log(2\pi\sigma^2_x)+\nonumber\\
    &+&\sum_{t_1}E_q\big[\log \frac{p(\hat{\by_1}^{t_1},\bz_1^{t_1}|s_1^{t_1})}{q(\bz_1^{t_1})}\big]+\sum_{i_2} E_q\big[\log \frac{p(\hat{\by_2}^{t_2},\bz_2^{t_2}|s_2^{t_2})}{q(\bz_2^{t_2})}\big]. 
\end{eqnarray}
This can be seen as solving a regularized matrix decomposition problem as by collecting $x_{t_1,t_2}$ into the observation matrix $\bX$ and all items' embedding vectors into matrices $\hat{\bY}_1$ and $\hat{\bY}_2$ the first term (first line) above can be written as

\begin{equation}
    T_1=-\frac{1}{2\sigma_y^2}||\bX -\hat{\bY}_1^T \hat{\bY}_2||_F^2-\frac{n_1 n_2}{2}\log(2\pi\sigma^2_x),
\end{equation}
while the second term (line) serves as a regularization term. The regularization term is in fact consisting of two models, one for each type of item, of the form studied in the previous sections. When the observation matrix is very large and also has low intrinsic dimension, as is typically the case in collaborative filtering applications, the first term can be fit very well resulting in low $\sigma_x^2$, and making the regularization term adapting to the natural data embedding. Therefore, as our goal for the two $p(\by,\bz|s)$ components is not to interfere with natural embeddings of the data, but rather to {\bf explain} the statistical patterns so that they would transfer to new semantic items with new descriptions and previously unseen interaction patterns in testing, we can fit the first term independently of the other two, after which the joint modeling problem separates into two separate ones, e.g. extracting the plant features that explain their embeddings and extracting the geographic areas features to explain their own embeddings.

For a given embedding dimension $n_e$ the term $T_1$ is minimized when $\hat{\bY}_1$ and $\hat{\bY}_2$ are related to the SVD of $\bX$, which reduces the dimensionality to $n_e$ $\bX\approx \bU \bS \bV^T$, where $n_1\times n_e$ matrix $U$ and the $n_2 \times n_e$ matrix $V$ are orthonormal and $S$ is a diagonal matrix. For example, the term is minimized when 
 \begin{equation}
     \hat{\bY_1}=\bU \bS^{\alpha} \bR, \quad \hat{\bY_2}=\bR^T \bS^{1-\alpha} \bV^T,
     \label{eq:decomposition}
 \end{equation}
 for arbitrary rotation matrix $\bR$ and a scalar $\alpha\in[0,1]$.

Therefore, if we solve the matrix decomposition problem in $T_1$ we can then fit the features for the two sets of items separately as the second term is decomposed:
\begin{equation}
    T_2=\sum_{t_1}E_q\big[\log \frac{p(\hat{\by_1}^{t_1},\bz_1^{t_1}|s_1^{t_1})}{q(\bz_1^{t_1})}\big]+\sum_{i_2} E_q\big[\log \frac{p(\hat{\by_2}^{t_2},\bz_2^{t_2}|s_2^{t_2})}{q(\bz_2^{t_2})}\big].
\end{equation}

In other words, by computing the embeddings using a standard matrix decomposition technique, we can build models that generate these item embeddings $\by$ using the semantics of the items $s$.

In case of movie preference datasets, like the Netflix Prize dataset, and data from many other recommender systems,  the second set of items -- users -- are usually not described semantically (through information like their demographics, their writing, etc.). Privacy reasons are often quoted for this, but it is also true that people's complex overall behavior, as it relates to the task, is better represented simply by their preferences. I.e. knowing which movies they have watched is a better indicator of whether or not they will like a particular new movie for example, than is some description of their life story, even if it were available. In other words, in our Netflix prize experiments, we only fit the movie embeddings.

\end{document}